\pdfoutput=1
\documentclass{article}

\usepackage[preprint]{neurips_2024}
\usepackage{CJKutf8}

\usepackage[utf8]{inputenc} % allow utf-8 input
\usepackage[T1]{fontenc}    % use 8-bit T1 fonts
\usepackage{hyperref}       % hyperlinks
\usepackage{url}            % simple URL typesetting
\usepackage{tabularx}
\usepackage{colortbl}
\usepackage{arydshln}
\usepackage{booktabs}       % professional-quality tables
\usepackage{amsfonts}       % blackboard math symbols
\usepackage{nicefrac}       % compact symbols for 1/2, etc.
\usepackage{microtype}      % microtypography
\usepackage{xcolor}         % colors
\usepackage{graphicx}
\usepackage{arydshln}
\usepackage{wrapfig}
\usepackage{amsmath}
\usepackage{natbib}
\setcitestyle{numbers,square,comma}
\usepackage{enumitem}
\usepackage{pifont}

\definecolor{dark2green}{rgb}{0.1, 0.65, 0.3}

\newcommand{\first}[1]{\textbf{\textcolor{dark2green}{#1}}}
\newcommand{\second}[1]{\textbf{\textcolor{black}{#1}}}
\DeclareUnicodeCharacter{2212}{\ensuremath{-}}

\title{Motif-driven Subgraph Structure Learning\\ for Graph Classification}

\author{%
    Zhiyao Zhou$^{1}$, Sheng Zhou$^{2}$\thanks{Corresponding Author}, Bochao Mao$^{2}$, Jiawei Chen$^{1}$\\
  \textbf{Qingyun Sun}$^{3}$ \textbf{Yan Feng}$^{1}$, \textbf{Chun Chen}$^{1}$, \textbf{Can Wang}$^{1}$\\
  $^{1}$College of Computer Science, Zhejiang University, Hangzhou, China \\
  $^{2}$School of Software Technology, Zhejiang University, Ningbo, China\\
  $^{3}$School of Computer Science and Engineering, Beihang University\\
}

\begin{document}

\maketitle

\begin{abstract}
  To mitigate the suboptimal nature of graph structure, Graph Structure Learning (GSL) has emerged as a promising approach to improve graph structure and boost performance in downstream tasks. 
  Despite the proposal of numerous GSL methods, the progresses in this field mostly concentrated on node-level tasks, while graph-level tasks (e.g., graph classification) remain largely unexplored. 
  Notably, applying node-level GSL to graph classification is non-trivial due to the lack of find-grained guidance for intricate structure learning. 
  Inspired by the vital role of subgraph in graph classification, in this paper we explore the potential of subgraph structure learning for graph classification by tackling the challenges of key subgraph selection and structure optimization.
  We propose a novel \textit{Motif-driven Subgraph Structure Learning} method for Graph Classification (MOSGSL). Specifically, MOSGSL incorporates a subgraph structure learning module which can adaptively select important subgraphs. A motif-driven structure guidance module is further introduced to capture key subgraph-level structural patterns (motifs) and facilitate personalized structure learning. Extensive experiments demonstrate a significant and consistent improvement over baselines, as well as its flexibility and generalizability for various backbones and learning procedures. The code is available in the \url{https://github.com/OpenGSL/OpenGSL}.
\end{abstract}

\section{Introduction}

As the primary approach for representation learning on graphs, Graph Neural Networks (GNNs) have been successfully applied to various tasks such as node classification~\cite{kipf2017semisupervised, graphsage, velivckovicgraph}, link prediction~\cite{zhang2018link, cai2021line}, and graph classification~\cite{xupowerful, you2020graph}. However, extensive studies have shown that GNNs are largely constrained by the quality of the input graph structure~\cite{dai2018adversarial, zugner_adversarial_2019}. Real-world data often inevitably exhibit suboptimal characteristics (e.g., missing connections during data collection or spurious links from malicious attacks), which significantly deteriorates the predictive power of GNN models~\cite{zhu2019robust}. 

\textit{Graph Structure Learning (GSL)}, a concept proposed to address potential flaws in the original graph structures from a \textit{data-centric}~\cite{zha2023data-centric-perspectives, zhou2023opengsl} perspective, is gaining increasing attention. 
Considerable GSL methods~\cite{liu2022towards, jin2020graph, wang2021graph, yu2021graph} have been proposed to improve the quality of graph structure and further boost the performance of downstream task, under the guidance of node labels or pretext tasks.
% Notably, recent works~\cite{fatemi2023ugsl, li_gslb_2023} have found that most GSL methods can be reformulated into a unified paradigm which learns the structure under the guidance of node labels or pretext tasks. 
% While such GSL paradigm has been widely shown to successfully boost performance in \textit{node-level tasks} (mostly node classification)~\cite{zhou2023opengsl, li_gslb_2023}, it is not inherently suitable for \textit{graph-level tasks} (e.g., graph classification), which remain largely unexplored in the GSL literature. 
Despite the success of GSL in \textit{node-level tasks} (e.g., node classification)~\cite{zhou2023opengsl, li_gslb_2023}, the potential of GSL in \textit{graph-level tasks} (e.g., graph classification) remain largely unexplored in the literature.
Specifically, graph classification has been shown extremely vulnerable to suboptimal graph structures~\cite{ijcai2023p449, xia2024gnncert}.
This notable gap significantly impedes the applicability of GSL in numerous real-world scenarios, including social group identification~\cite{10.1145/3038912.3052588, morris2020tudataset}, protein property prediction~\cite{10.1093/bioinformatics/bti1007, gligorijevic2021structure} and so on.

Although important, extending node-level GSL methods to graph-level tasks is a non-trivial problem.
\begin{wrapfigure}{r}{0.55\linewidth}
\label{fig:label}
\vspace{-1mm}
    \centering
    \includegraphics[width=\linewidth]{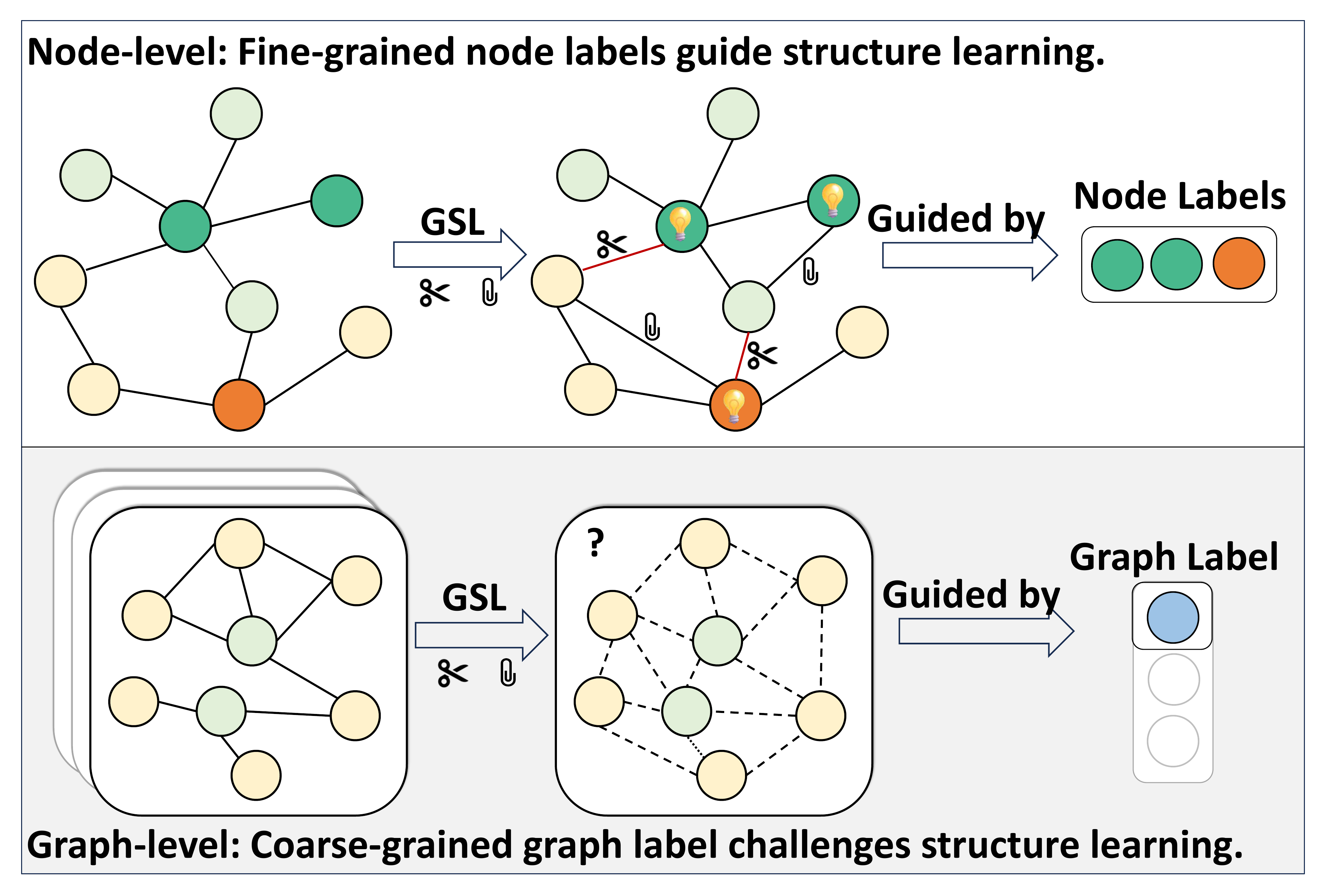}
    \vspace{-5mm}
    \caption{The problem of coarse-grained supervision in graph classification.}
    \vspace{-5mm}
\end{wrapfigure}
Concretely, node-level GSL methods mostly rely on fine-grained node labels to guide node representation learning and promote intricate structure learning~\cite{fatemi2021slaps}. 
However, in graph-level tasks where only single label is available for the entire graph, the guidance for intricate structure learning is insufficient, as shown in Figure~\ref{fig:label}.
Meanwhile, subgraphs have been widely recognized vital in graph-level tasks~\cite{0eb15824-3d89-3bf3-8d79-2cf60ef065fa, Girvan_2002}, characterizing discriminative information at a higher level than individual nodes/edges. 
This raises an interesting research question: \textit{Can we fully exploit subgraphs and coarse-grained graph label to guide structure learning in graph-level tasks?}

% 之前方法的成功大多依赖于细粒度的节点标签来精细化的指导节点表征和图结构的更新, 但是这在图级别的任务中由于我们只能获得整图的标签，因此我们无法获得足够的指导来进行精细化的结构更新。
% 与此同时，之前的图级别任务中已经表明子图相对于节点对于刻画图的全局特性具有更重要的意义，因此， 我们尝试回答将的一个问题， 我们是否可以在突击别的任务中利用子图和有限的全局标签来引导结构学习？

Tackling this problem meets two significant challenges: 1) Each graph may comprise a large number of subgraphs, yet only a finite subset is valuable for the final graph-level prediction~\cite{sun2021sugar}. \textit{How to accurately select key subgraphs from a vast pool while optimizing their structures to aid in the graph-level tasks?} 
2) Subgraphs from graphs of the same category often exhibit discriminative structural patterns~\cite{peng2020motif} (referred to as ``motif'' in this work), which is valuable for graph-level tasks. \textit{How to effectively capture representative motifs of each graph category and guide the structure learning of subgraphs?}

In this paper, we focus on the novel yet nontrivial problem of applying GSL to graph-level task (i.e., graph classification). In light of the above challenges, we go beyond the conventional GSL paradigm and present a novel \textbf{MO}tif-driven \textbf{S}ub\textbf{G}raph \textbf{S}tructure \textbf{L}earning method for Graph Classification (abbreviated as \textbf{MOSGSL}). 
To solve the first challenge, we design a \textit{subgraph structure learning module} (SGSL) which conducts structure learning on each subgraph independently. 
A gate mechanism is introduced to adaptively combine the refined substructures according to their estimated importance scores. In this way, SGSL can accurately select key subgraphs and meanwhile derives refined subtructures for them. To solve the second challenge, we propose \textit{a motif-driven structure guidance module} comprising two phases: motif extraction and subgraph-motif alignment. In the motif extraction phase, motifs are updated based on refined substructures from key subgraphs in a label-wise manner, which well captures representative patterns for each category. In the motif alignment phase, we derive a contrastive loss that dynamically guide structure learning on each subgraph towards its corresponding target (motif). Notably, these two steps are iteratively conducted to ensure mutual enhancement between structure learning and motif, thus facilitating precise and personalized structure learning under coarse-grained supervision.

% to alleviate coarse-grained supervision and provide diversified guidance for structure learning, we further propose a motif-driven structure guidance module, where "motif" refers to extracted crucial structural patterns. Specifically, in the motif extraction phase, motifs are updated based on the distribution of learned substructures. Then in the motif alignment phase, we derive a contrastive loss that dynamically guide each subgraph towards its corresponding target (motif). Notably, these two steps are iteratively conducted to ensure mutual enhancement between structure learning and motif extraction, facilitating precise and personalized structure learning for important subgraphs.

Extensive experiments on five datasets have demonstrated the significant advantages of MOSGSL compared with other baselines. Moreover, MOSGSL is shown broadly effective equipped with various backbones. Furthermore, we validate the versatility of MOSGSL in different learning procedures, including preprocessing, co-training, and test-time settings, showcasing its potential for various real-world applications. 

To sum up, the main contributions of our work can be summarized as follows:

\begin{enumerate}
    \item We study a novel yet nontrivial problem of applying GSL to graph classification and provide new insights on the problem of coarse-grained supervision, as well as identify unique challenges of key subgraph selection and structure optimization.
    \item To tackle these challenges, we go beyond the conventional GSL paradigm and propose the MOSGGL framework. MOSGSL conducts structure learning at the subgraph level and can perceive the importance of subgraphs. By incorporating a motif-driven structure guidance module, the structural patterns of important subgraphs are further captured as motifs, which guide precise and personalized structure learning under coarse-grained global labels.

\item Comprehensive experiments across multiple datasets demonstrate the efficacy of our proposed MOSGSL. MOSGSL also allows flexible interchange of backbones and can adapt to different learning procedures.
\end{enumerate}

\section{Related Work}

\paragraph{Graph Neural Networks.} Graph Neural Networks (GNNs) have become the mainstream approach for learning on graph-structured data~\cite{wu2020comprehensive}. The superiority of GNNs can be attributed to the ability to leverage both semantic and topological information via a message-passing scheme~\cite{kipf2017semisupervised}. Numerous endeavors~\cite{chienadaptive, xupowerful, xu2018representation, velivckovicgraph, gasteigerpredict} have been made to propose more advanced GNNs, which have achieved state-of-the-art performance on various tasks (e.g., node classification, graph classification). However, extensive studies have identified vulnerabilities of GNNs when faced with suboptimal structure~\cite{dai2018adversarial, zugner_adversarial_2019}. Relying heavily on the quality of the input graph structure, GNNs exhibit limited predictive power when encountering spurious connections or absent links within the structure, which are inevitably present in real-world data~\cite{zhu2019robust}.

\paragraph{Graph Structure Learning.} Recent years have witnessed a growing interest in graph structure learning~\cite{in2024self, duan2024structural, zou2023se, zhao2023self, wang2023prose, liu2022towards, zhao2021data, yu2021graph}, a concept proposed to address potential flaws in graph structures from a data-centric perspective. These methods aim to enhance the quality of graph structures to boost the performance of downstream tasks. Despite the plethora of GSL methods, most existing studies focus on node-level tasks, while graph-level tasks where we need to learn structures based on global signals, remain largely unexplored. 

Some unsupervised GSL methods~\cite{liu2022towards, li2022reliable} that modify structures in the data preprocessing stage seem applicable to graph-level tasks. However, they rely on pretext tasks that highlight differences between individual nodes and are inherently unsuitable for graph classification. VIBGSL~\cite{sun2022graph} learns the optimal graph structure for multiple graphs with the assistance of Information Bottleneck (IB) theory. As a simple extension of the conventional GSL paradigm, it still faces challenges in refining structures accurately for subgraphs under coarse-grained supervision. HGP-SL~\cite{zhang2019hierarchical} also conducts structure learning at the subgraph level. In comparison to HGP-SL, our method differs by learning substructures simultaneously rather than serially and further introducing motifs as auxiliary guidance. Additionally, a recent work~\cite{wang2023prose} also proposes subgraph-level structure learning, however, it only targets node classification without considering graph-level tasks.

\section{Methodology}
\label{sec:method}

\begin{figure}[t]
  \centering
  \includegraphics[width=\textwidth]{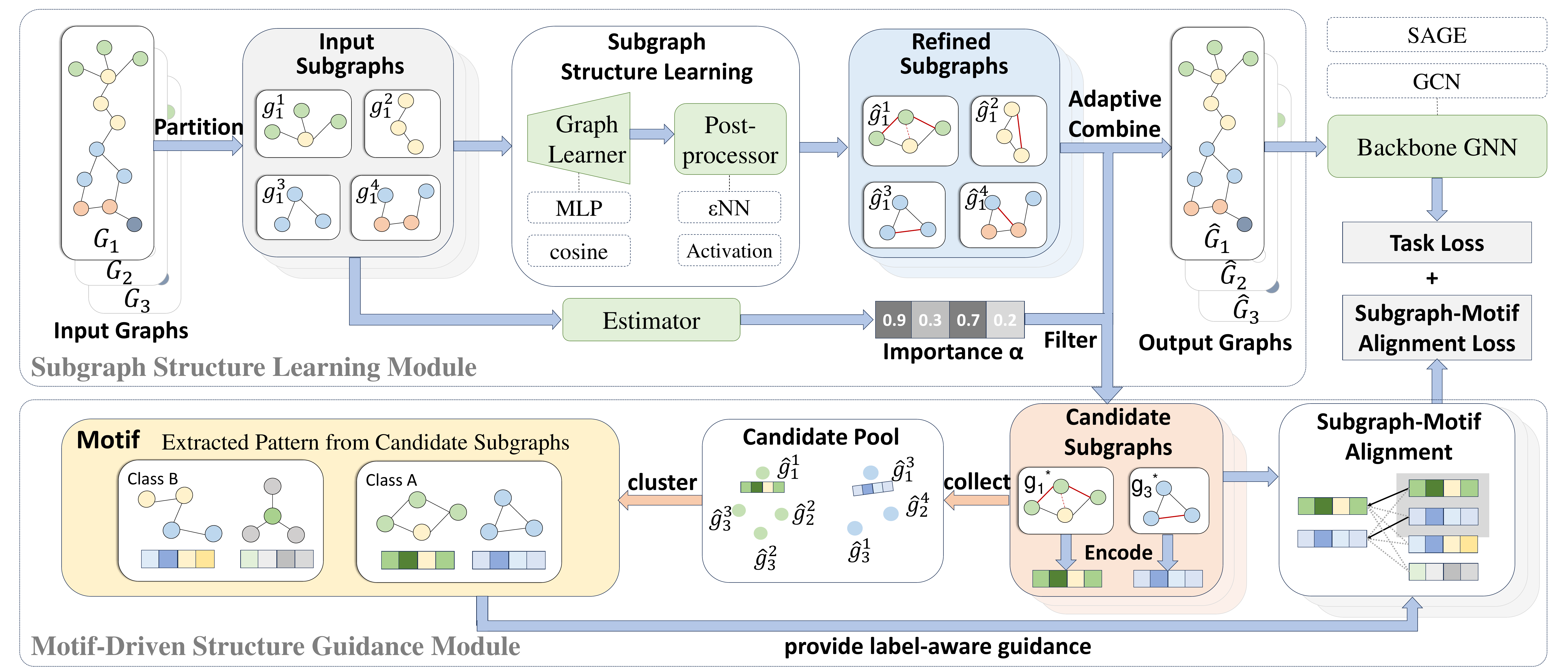}
  \caption{The framework of proposed MOSGSL, including a Subgraph Structure Learning Module (SGSL) and a Motif-driven Structure Guidance Module. In SGSL, we conduct structure learning on subgraphs separately and use a gate mechanism to adaptively combine the learned substructures. Important subgraphs are then filtered as candidate subgraphs, based on which motif alignment and motif extraction are iteratively conducted. Several components can be flexibly interchanged in MOSGSL.}
  \label{fig:model}
  \vspace{-4pt}
\end{figure}

In this section, we elaborate on the \textbf{MO}tif-Driven \textbf{S}ub\textbf{G}raph \textbf{S}tructure \textbf{L}earning method designed specifically for graph-level tasks, referred to as MOSGSL. The overall framework is depicted in Figure~\ref{fig:model}. We first provide necessary preliminaries concerning GSL in graph classification, introduce the conventional graph structure learning pipeline, and highlight its constraints in graph classification. Subsequently, we expand upon this pipeline to the subgraph level, introducing a subgraph structure learning module that stands out by identifying and emphasizing important subgraphs. Finally, to provide auxiliary guidance under coarse-grained global signals, we design a motif-driven structure guidance module that captures diverse structural patterns as motifs and guides towards precise and personalized structure learning for subgraphs.

\subsection{Preliminaries}

\paragraph{Notations.} In this work we focus on the graph classification task. Given a set of labeled graphs $\mathcal{D}=\{(G_1,y_1),(G_2,y_2),...\}$ where $y_i \in \mathcal{Y}$ is the label of graph $G_i \in \mathcal{G}$, the goal of graph classification is to learn a mapping $f:\mathcal{G}\rightarrow \mathcal{Y}$ that maps graphs to the set of labels. $f$ is typically parameterized as a GNN model. Each graph $G_i$ containing $n_i$ nodes can be represented as $(A_i,X_i)$, where $A_i \in \{0,1\}^{n_i \times n_i}$ is the adjacency matrix and $X_i \in \mathbb{R}^{n_i \times F}$ denotes node features with $F$ dimensions. 

% With a slight abuse of notation, we omit superscripts $i$ when describing operations on a single graph in the following text.

\label{sec31}
\paragraph{Conventional Graph Structure Learning Paradigm.}
Numerous Graph Structure Learning (GSL) methods have been proposed for node-level tasks. Despite their diverse designs and implementations, recent works~\cite{fatemi2023ugsl, li_gslb_2023} have found that most of them can be reformulated into a unified paradigm, which includes a Graph Learner and a Processor.:
\begin{equation}
    \hat{A}_i = \text{GSL}(G_i)=\text{Processor} \circ  \text{GraphLearner}(G_i).
\end{equation}
Specifically, the GraphLearner is used to compute pair-wise edge scores, producing a similarity matrix. Subsequently, the learned adjacency matrix is further refined by the Processor. Commonly used techniques include k-nearest neighbors (KNN), $\epsilon$-nearest neighbors ($\epsilon$-NN), concrete relaxation, symmetrization, etc. Finally, the learned structure is fed into the backbone GNN $f$ to learn representations for downstream tasks (here we consider graph-level task):
\begin{equation}
    \min \mathcal{L}(f(\hat{A}_i,X_i),y_i).
\end{equation}
As discussed earlier, such a framework may not be inherently well-suited for graph-level tasks due to coarse-grained supervision and inadequate emphasis on subgraphs. Given these considerations, our aim is to go beyond the conventional graph structure learning paradigm and design a general GSL framework targeted for graph classification that can: 1) accurately select key subgraphs for each graph and refine their substructures and 2) effectively capture distinct subgraph patterns across various categories and in turn provide auxiliary guidance for structure learning.

\subsection{Subgraph Structure Learning Module}

Here we introduce a \textbf{S}ub\textbf{G}raph \textbf{S}tructure \textbf{L}earning module (SGSL) as an extension of the conventional GSL paradigm. We first partition the graph into several subgraphs and estimate their importance. Afterwards, we conduct structure learning on these subgraphs respectively, and the learned substructures are adaptively integrated according to their importance.

Various techniques can be employed to partition a graph into $K$ subgraphs $G_{i,\text{sub}}=\{g_i^k|k=1,2...,K\}$. Following~\cite{sun2021sugar}, we opt for the simple breadth-first search (BFS) algorithm. For a given graph $G_i=(A_i,X_i)$, we first sort the nodes based on their degrees and designate the top $K$ nodes as the central nodes for the subgraphs. We then utilize BFS to extract a subgraph for each central node, with maximum number of nodes in each subgraph limited to $M$.  The specific values of $K$ and $M$ are determined based on the scale of the datasets to ensure appropriate coverage on the original graph. 

After obtaining a series of subgraphs $G_{i,\text{sub}}=\{g_i^k|k=1,2...,K\}$, we need to estimate their respective importance scores. Here we use a GNN-based encoder to obtain representations for each subgraph:
\begin{equation}
    z_i^k=\text{Pooling}(H_i^k),H_i^k=\text{Encoder}(g_i^k),\forall k \in \{1,2,...,K\}.
\end{equation}
The importance score $\alpha_i^k$ for each subgraph is calculated as $\alpha_i^k=\frac{z_i^kp}{||p||}$, where $p$ represents a learnable vector. Meanwhile, we perform structure learning on these subgraphs to obtain refined graph structures $\hat{G}_{i,\text{sub}}=\{ \hat{g}_i^k \}$, where each refined subgraph $\hat{g}_i^k$ can be learned as in Section~\ref{sec31}:
\begin{equation}
    g_i^k = (A_i^k, X_i^k)=\text{GSL}(g_i^k),\forall k \in \{1,2,...,K\}.
\end{equation}
In the above process, we separate different subgraphs within a graph and conduct structure learning on them separately. In this way, we treat each subgraph as an individual entity and focus only on the local characteristics of the subgraphs during structure learning for each subgraph, thus avoiding interference from other irrelevant parts. Then with the learned substructures, we adopt a gate mechanism and adaptively combine them to obtain the final optimal graph structure:
\begin{equation}
    \hat{A}_i = \gamma\sum_{k=1}^K \alpha_i^k\hat{A}_i^k+(1-\gamma)A_i,
\end{equation}
where $\gamma$ controls the proportion of integration with the original graph structure. Unlike conventional GSL methods that treat the entire graph uniformly, SGSL divides the graph into several subgraphs, learns structures for subgraphs independently, and utilizes importance scores as weights when fusing substructures. By introducing such a gate mechanism for subgraphs, SGSL has the ability to perceive the importance of subgraphs during structure learning and mitigates the influence of irrelevant subgraphs, leading to more stable optimization under coarse-grained signals. Besides, the partitioning approach we previously used allows edges to be shared by multiple subgraphs. In this case, the fusion process can also be viewed as a weighted vote, where the final edge weight is determined by more important subgraphs it belongs.

Lastly, with the refined subgraphs $\hat{G}_{i,\text{sub}}$ and estimated importance scores $\{\alpha_i^k|k=1,2...,K\}$, we select subgraphs with a high importance score to obtain candidate subgraphs $\hat{G}_{\text{can}}$:
\begin{equation}
\hat{G}_{i,\text{can}}=\text{Top}\epsilon(\{(\alpha_i^k, \hat{g}_i^k)|k=1,2...,K\}).
\end{equation}

Here we select the top $\epsilon$ proportion of subgraphs in terms of their importance scores. Based on these candidates, we extract representative patterns (i.e., motifs) within them and further guide the learned substructures towards diversified targets in the following module.

\subsection{Motif-driven Structure Guidance Module}

We have now introduced the SGSL module, which conducts structure learning at the subgraph level and adaptively selects key subgraphs. Intuitively, subgraphs from graphs of different classes exhibit distinct structural patterns, which can be viewed as the targets for structure learning on subgraphs. Besides, different subgraphs within a single graph may also have varying learning targets. Based on the intuitions, referring to the extracted subgraph-level structural pattern as ``motif'', we propose the motif-driven structure guidance module to guide the model towards more precise and personalized structure learning under coarse-grained supervision. 

Specifically, based on the candidate subgraphs selected above, we iteratively perform subgraph-motif alignment and motif extraction. For each minibatch, these candidate subgraphs will be mapped to motifs belonging to the corresponding class, evolving into a subgraph-motif alignment loss. Meanwhile, the candidate subgraphs are collected across multiple minibatches in an epoch and are clustered to update motifs in a class-wise manner.

\paragraph{Subgraph-Motif Alignment Phase.} Assuming we have obtained a series of motifs capturing essential structural patterns for subgraphs across different classes $\mathcal{M}=\{m^j|j\in\{1,2,...,L\}\}$ with $\mathrm{y}(\cdot):\{1,2,...,L\}\rightarrow \mathcal{Y}$ indicating the class to which each motif belongs, the challenge lies in mapping the learned substructures $\hat{G}_{i,\text{can}}$ with the corresponding motifs. Since the graph matching problem is computationally intractable, we opt for modeling motifs from a representational perspective. Namely, we represent motifs as $\mathcal{M}=\{m^j|m^j \in R^{d\times 1},j\in\{1,2,...,L\}\}$. Therefore, computing the match ratio between learned substructures and motifs becomes straightforward, which can be formalized as:
\begin{equation}
    s_i^{k,j}=\phi(\text{Encoder}(\hat{g}_i^k), m^j),
\end{equation}
where the Encoder is a GNN-based encoder, and $\phi(\cdot, \cdot):R^{d\times 1}\times R^{d\times 1}\rightarrow R$ is a function that calculates the similarity between two representations (e.g., euclidean distance, cosine similarity). Treating motifs of the corresponding class as positive samples and motifs of other classes as negative samples, we develop a contrastive loss that enforces the learned substructures to align with specific motifs:
\begin{equation}
    \mathcal{L}_{\text{motif}}(\hat{G}_{i,\text{can}}, \mathcal{M})=\frac{1}{|\hat{G}_{i,\text{can}}|}\sum_{\hat{g}_i^k \in \hat{G}_{i,\text{can}}}\log \frac{\min_{\mathrm{y}(j)=y_i}e^{s_i^{k,j} / t}}{\sum_{\mathrm{y}(j)\neq y_i} e^{s_i^{k,j}/ t}},
\end{equation}
where $t$ denotes temperature. It's worth noting that the motif alignment loss provides diverse learning targets not only for subgraphs from different categories via the label-aware motif matching, but also for subgraphs within the same graph via the use of $\min$ instead of $\text{sum}$ in the numerator. With this motif alignment loss, we aim to guide structure learning on candidate subgraphs towards diversified learning objectives (i.e., motifs), thus promoting personalized structure learning in an explicit manner. At the same time, serving as common patterns extracted from the data, motifs also function as auxiliary guidance that aids in more precise structure learning under coarse-grained supervision.

\label{sec33}
\paragraph{Motif Extraction Phase.}
The remaining problem is how to effectively extract key patterns as motifs from $\mathcal{\hat{G}}_{\text{can}}=\bigcup_i \hat{G}_{i,\text{can}}$, referring to candidate subgraphs collected from the training graphs in an epoch. To achieve this, we decide to cluster candidate subgraphs in a class-wise style and use the centroids to update representations of motifs belong to the corresponding class.

Specifically, we update motif representations using k-means algorithm for simplicity. Within each class, we employ the original motifs $\mathcal{M}^c=\{m^j|\mathrm{y}(j)=c\}$ as initial centroids and consolidate the representations of corresponding subgraphs via k-means:
\begin{equation}
    M^c_{new}=\text{KMeans}(\{\text{Encoder}(\hat{g}_i^k)|\hat{g}_i^k \in \mathcal{\hat{G}}_{\text{can}}, y_i=c\}), \text{started from } M^c_{old}, \forall c \in \mathcal{Y}.
\end{equation}
In this process, candidate subgraphs from graphs of a specific class are used as the source for extracting corresponding motifs. This class-wise approach ensures the correspondence between motifs and categories and explicitly distinguishes motifs across different classes. Additionally, we use motifs from the last round as initial centroids to keep the orderliness among motifs unchanged in the motif extraction phase.

\subsection{Proposed MOSGSL}
In summary, our proposed MOSGSL conducts structure learning at the subgraph level identifies important subgraphs via the SGSL module, and further incorporates a motif-driven structure guidance module to capture common subgraph patterns and use them to guide personalized structure learning, respectively realized in the motif extraction phase and subgraph-motif alignment phase. Notably, motif extraction and subgraph-motif alignment work in an iterative manner, akin to~\cite{zou2023se, wang2021graph}. In each epoch, the model is trained in an end-to-end manner by optimizing the subgraph-motif alignment loss and the task loss:
\begin{equation}
    \min \mathcal{L}(f(G_i),y_i)-\lambda \mathcal{L_{\text{motif}}}(\hat{G}_{i,\text{can}}, \mathcal{M}).
\end{equation}
Then every $\eta$ epoch we update the motif using collected candidate subgraphs. This iterative framework bolsters the mutual enhancement between structure learning and motif extraction, i.e., better structures lead to better motifs, and better motifs in turn guide better structure learning. Some experimental results supporting this intuition can be found in Sec~\ref{dynamic}. Additionally, in terms of motif initialization, we provide two options: random initialization and initialization from pre-trained representations. (See Appx~\ref{motif} for more details.) MOSGSL is also a general and flexible framework, where several components can be freely interchangeable, as shown in Figure~\ref{fig:model}.

\section{Experiments}
\label{sec:experiments}

\subsection{Experimental Settings}

\paragraph{Datasets.} We performed extensive experiments on five datasets commonly used in the literature~\cite{morris2020tudataset, errica_fair_2020}, including three social datasets (IMDB-B, IMDB-M, and RDT-B) and two biological datasets (ENZYMES and PROTEINS). These datasets are from diverse domains with distinct characteristics, ensuring a thorough evaluation of MOSGSL's efficacy. Additional information about these datasets can be found in Appx~\ref{dataset}.

\paragraph{Baselines.} We encompass a range of cutting-edge methods as baselines, which can be divided into three groups. The first group includes several classic GNNs: GCN~\cite{kipf2017semisupervised}, GraphSAGE~\cite{graphsage}, and GIN~\cite{xupowerful}. The second group comprises GSL methods that can learn structures for graph-level tasks, including HGP-SL~\cite{zhang2019hierarchical} and VIBGSL~\cite{sun2022graph}, which are the only two methods considering graph-level tasks in their original papers. In addition, for a more comprehensive comparison, we further incorporate some GSL methods (i.e., GRCN~\cite{yu2021graph}, GAUG~\cite{zhao2021data}, NeuralSparse~\cite{zheng2020robust}) that can be extended to graph-level tasks with minor adjustments. Finally, since we target graph-level tasks, methods in the last group include several typical graph-pooling methods~\cite{baek2021accurate, gao2019graph, diehl2019edge}.

% which also serve as foundational backbones for more complicated frameworks in subsequent groups. 

\paragraph{Evaluation Protocol and Hyper-parameters.} Following many previous works~\cite{sun2022graph, errica_fair_2020, li_gslb_2023}, we perform 10-fold cross-validation and report the average accuracy along with the standard deviation. We carefully replicate all baseline algorithms in accordance with their original papers. Common hyperparameters, such as learning rate ($lr\in \{0.1,0.01,0.001\}$), weight decay ($wd \in \{5\mathrm{e}{-4},5\mathrm{e}{−5},5\mathrm{e}{−6},0\}$), and batch size ($bs\in \{32,64,128\}$), are carefully selected according to the performance on validation set. We train all methods for 400 epochs and early stopping strategy is adopted, where training is stopped if the validation loss does not decrease for 50 consecutive epochs. We use the Adam optimizer and L2 regularization for all methods. More details regarding the experimental procedures and hyperparameter configurations can be referenced in Appdx~\ref{hyper}.

\begin{table}[!t]
\renewcommand\arraystretch{1.05}
    \caption{Performance (\%) of all models. Shown is the mean~±~s.d.~of 10 runs on different folds. Highlighted are the top \first{first} and \second{second}. "OOM" denotes out of memory.}
    \label{tab:1}
    \centering
    % \small
    \fontsize{9pt}{9pt}\selectfont
    \setlength\tabcolsep{6pt} % default value: 6pt
    \begin{tabular}{lcccccc}\toprule
    \textbf{Model} &\textbf{IMDB-B} &\textbf{IMDB-M} &\textbf{RDT-B} &\textbf{ENZYMES} &\textbf{PROTEINS} &\textbf{Avg. Rank} \\
    \midrule
GCN & 72.60{\tiny$\pm$3.53} & 49.60{\tiny$\pm$1.94} & 70.60{\tiny$\pm$3.69} & 48.67{\tiny$\pm$5.71} & 66.49{\tiny$\pm$4.58} & 8 \\
SAGE & 72.80{\tiny$\pm$2.48} & 49.60{\tiny$\pm$2.39} & 67.80{\tiny$\pm$4.24} & 55.50{\tiny$\pm$5.11} & 68.56{\tiny$\pm$4.93} & 7 \\
GIN & 72.00{\tiny$\pm$2.37} & 48.33{\tiny$\pm$2.05} & 75.50{\tiny$\pm$2.40} & 54.00{\tiny$\pm$5.17} & 69.18{\tiny$\pm$2.85} & 8 \\
% \noalign{\vskip 1.5pt}
\midrule
% \noalign{\vskip 1.5pt}
GMT & 73.30{\tiny$\pm$3.16} & 50.07{\tiny$\pm$4.46} & 84.15{\tiny$\pm$2.47} & \second{60.33{\tiny$\pm$6.61}} & \second{73.77{\tiny$\pm$5.69}} & 3.2 \\
TopKPool & \second{74.70{\tiny$\pm$2.67}} & 49.93{\tiny$\pm$3.13} & \second{84.80{\tiny$\pm$6.25}} & 52.67{\tiny$\pm$7.54} & 71.33{\tiny$\pm$6.56} & 4.2 \\ 
EdgePool & 73.40{\tiny$\pm$3.63} & \second{50.80{\tiny$\pm$3.29}} & 81.15{\tiny$\pm$3.56} & 55.17{\tiny$\pm$9.26} & 71.34{\tiny$\pm$4.55} & 3.8 \\
HGP-SL & 72.90{\tiny$\pm$3.05} & 48.93{\tiny$\pm$1.82} & OOM & 49.67{\tiny$\pm$9.42} & 71.88{\tiny$\pm$4.35} & 7.4 \\
\midrule
GRCN & 72.50{\tiny$\pm$2.64} & 49.47{\tiny$\pm$2.35} & OOM & 46.83{\tiny$\pm$7.76} & 60.69{\tiny$\pm$2.29} & 9.8 \\
GAUG & 70.90{\tiny$\pm$3.54} & 49.80{\tiny$\pm$3.63} & OOM & 41.17{\tiny$\pm$7.58} & 59.57{\tiny$\pm$7.00} & 10.4 \\ 
NeuralSparse & 72.90{\tiny$\pm$3.75} & 50.47{\tiny$\pm$2.63} & 70.10{\tiny$\pm$2.79} & 45.33{\tiny$\pm$5.87} & 68.47{\tiny$\pm$3.85} & 7 \\ 
VIBGSL & 73.90{\tiny$\pm$2.47} & 49.33{\tiny$\pm$3.08} & 68.40{\tiny$\pm$3.81} & 42.50{\tiny$\pm$4.53} & 71.61{\tiny$\pm$4.57} & 7.2 \\
\midrule
MOSGSL & \first{75.50{\tiny$\pm$2.25}} & \first{51.60{\tiny$\pm$2.05}} & \first{87.45{\tiny$\pm$2.19}} & \first{63.50{\tiny$\pm$5.40}} & \first{74.03{\tiny$\pm$5.41}} & 1 \\
    \bottomrule
    \end{tabular}
    % \vspace{-1mm}
\end{table}

\begin{table}[!t]
\vspace{-2mm}
\renewcommand\arraystretch{1.05}
    \caption{Performance (\%) of all GSL models using various backbones. Shown is the mean~±~s.d.~of 10 runs on different folds. Highlighted are the top \first{first} and \second{second}. "OOM" denotes out of memory.}
    \label{tab:2}
    \centering
    % \small
    \fontsize{9pt}{9pt}\selectfont
    \setlength\tabcolsep{6pt} % default value: 6pt
    \begin{tabular}{lcccccc}\toprule
    \textbf{Model} &\textbf{IMDB-B} &\textbf{IMDB-M} &\textbf{RDT-B} &\textbf{ENZYMES} &\textbf{PROTEINS} &\textbf{Avg. Rank} \\
    \midrule
    \rowcolor{gray!12}\multicolumn{7}{c}{\textsc{Using GCN as Backbone}} \\
    \midrule
GCN & 72.60{\tiny$\pm$3.53} & 49.60{\tiny$\pm$1.94} & \second{70.60{\tiny$\pm$3.69}} & 48.67{\tiny$\pm$5.71} & 66.49{\tiny$\pm$4.58} & 3.8 \\
% \noalign{\vskip 1.5pt}
% \hdashline
% \noalign{\vskip 1.5pt}
% GMT & 73.30{\tiny$\pm$3.16} & 50.07{\tiny$\pm$4.46} & 84.15{\tiny$\pm$2.47} & 60.33{\tiny$\pm$6.61} & 73.77{\tiny$\pm$5.69} & - \\
% TopKPool & 74.70{\tiny$\pm$2.67} & 49.93{\tiny$\pm$3.13} & 84.80{\tiny$\pm$6.25} & 52.67{\tiny$\pm$7.54} & 71.33{\tiny$\pm$6.56} & - \\ 
% EdgePool & 73.40{\tiny$\pm$3.63} & 50.80{\tiny$\pm$3.29} & 81.15{\tiny$\pm$3.56} & 55.17{\tiny$\pm$9.26} & 71.34{\tiny$\pm$4.55} & - \\
% \noalign{\vskip 1.5pt}
% \hdashline
% \noalign{\vskip 1.5pt}
GRCN & 72.50{\tiny$\pm$2.64} & 49.47{\tiny$\pm$2.35} & OOM & 46.83{\tiny$\pm$7.76} & 60.69{\tiny$\pm$2.29} & 5.2 \\
GAUG & 70.90{\tiny$\pm$3.54} & 49.80{\tiny$\pm$3.63} & OOM & 41.17{\tiny$\pm$7.58} & 59.57{\tiny$\pm$7.00} & 5.8 \\ 
NeuralSparse & 72.90{\tiny$\pm$3.75} & \second{50.47{\tiny$\pm$2.63}} & 70.10{\tiny$\pm$2.79} & 45.33{\tiny$\pm$5.87} & 68.47{\tiny$\pm$3.85} & 3.4 \\ 
VIBGSL & \second{73.90{\tiny$\pm$2.47}} & 49.33{\tiny$\pm$3.08} & 68.40{\tiny$\pm$3.81} & 42.50{\tiny$\pm$4.53} & 71.61{\tiny$\pm$4.57} & 4.2 \\
HGP-SL & 72.90{\tiny$\pm$3.05} & 48.93{\tiny$\pm$1.82} & OOM & \second{49.67{\tiny$\pm$9.42}} & \second{71.88{\tiny$\pm$4.35}} & 3.8 \\

% \noalign{\vskip 1.5pt}
% \hdashline
% \noalign{\vskip 1.5pt}
MOSGSL & \first{75.50{\tiny$\pm$2.25}} & \first{51.60{\tiny$\pm$2.05}} & \first{87.45{\tiny$\pm$2.19}} & \first{63.50{\tiny$\pm$5.40}} & \first{74.03{\tiny$\pm$5.41}} & 1 \\
% \noalign{\vskip -1.5pt}
\midrule
    \rowcolor{gray!12}\multicolumn{7}{c}{\textsc{Using SAGE as Backbone}} \\
    \midrule
% \noalign{\vskip -1.5pt}
SAGE & 72.80{\tiny$\pm$2.48} & 49.60{\tiny$\pm$2.39} & 67.80{\tiny$\pm$4.24} & 55.50{\tiny$\pm$5.11} & 68.56{\tiny$\pm$4.93} & 4.4 \\
% \noalign{\vskip 1.5pt}
% \hdashline
% \noalign{\vskip 1.5pt}
% GMT & 73.60{\tiny$\pm$3.44} & 50.20{\tiny$\pm$2.83} & 84.00{\tiny$\pm$4.30} & - & 72.60{\tiny$\pm$2.96} & - \\
% TopKPool & 72.00{\tiny$\pm$4.24} & 48.60{\tiny$\pm$3.32} & 67.30{\tiny$\pm$8.07} & - & - & - \\ 
% EdgePool & 74.00{\tiny$\pm$3.89} & 50.00{\tiny$\pm$2.67} & 73.30{\tiny$\pm$3.60} & - & 65.96{\tiny$\pm$6.20} & - \\
% HGP-SL & \second{73.80{\tiny$\pm$3.63}} & \second{50.33{\tiny$\pm$1.80}} & OOM & 52.17{\tiny$\pm$7.96} & \first{73.78{\tiny$\pm$3.73}} & 2.6 \\
% \noalign{\vskip 1.5pt}
% \hdashline
% \noalign{\vskip 1.5pt}
GRCN & 73.20{\tiny$\pm$3.08} & 49.47{\tiny$\pm$2.88} & OOM & 48.17{\tiny$\pm$5.58} & 70.69{\tiny$\pm$5.68} & 5.4 \\
GAUG & 72.40{\tiny$\pm$3.03} & 49.53{\tiny$\pm$2.25} & OOM & 52.33{\tiny$\pm$9.02} & 67.03{\tiny$\pm$6.93} & 5.8 \\ 
NeuralSparse & 72.50{\tiny$\pm$3.75} & 49.87{\tiny$\pm$2.75} & \second{68.05{\tiny$\pm$3.21}} & \second{59.67{\tiny$\pm$5.15}} & 66.76{\tiny$\pm$6.00} & 4.2 \\ 
VIBGSL & \second{73.80{\tiny$\pm$2.15}} & 49.93{\tiny$\pm$1.63} & 66.95{\tiny$\pm$4.19} & 58.33{\tiny$\pm$4.05} & 70.72{\tiny$\pm$4.33} & 3 \\
HGP-SL & \second{73.80{\tiny$\pm$3.63}} & \second{50.33{\tiny$\pm$1.80}} & OOM & 52.17{\tiny$\pm$7.96} & \first{73.78{\tiny$\pm$3.73}} & 3.2 \\

% \noalign{\vskip 1.5pt}
% \hdashline
% \noalign{\vskip 1.5pt}
MOSGSL & \first{75.90{\tiny$\pm$2.84}} & \first{51.87{\tiny$\pm$1.65}} & \first{79.20{\tiny$\pm$3.65}} & \first{65.17{\tiny$\pm$5.89}} & \second{72.42{\tiny$\pm$3.84}} & 1.2 \\
% \noalign{\vskip 1.5pt}
\midrule
    \rowcolor{gray!12}\multicolumn{7}{c}{\textsc{Using GIN as Backbone}} \\
    \midrule
% \noalign{\vskip 1.5pt}
GIN & 72.00{\tiny$\pm$2.37} & 48.33{\tiny$\pm$2.05} & 75.50{\tiny$\pm$2.40} & 54.00{\tiny$\pm$5.17} & 69.18{\tiny$\pm$2.85} & 4.6 \\
% \noalign{\vskip 1.5pt}
% \hdashline
% \noalign{\vskip 1.5pt}
% GMT & 74.60{\tiny$\pm$3.37} & 48.60{\tiny$\pm$3.64} & 78.25{\tiny$\pm$6.02} & - & 73.50{\tiny$\pm$6.13} & - \\ 
% TopKPool & 73.30{\tiny$\pm$2.45} & 49.47{\tiny$\pm$2.20} & 77.30{\tiny$\pm$5.20} & - & 70.63{\tiny$\pm$5.44} & - \\
% EdgePool & 72.70{\tiny$\pm$4.47} & 48.07{\tiny$\pm$3.35} & 85.85{\tiny$\pm$5.00} & - & 68.64{\tiny$\pm$4.92} & - \\ 
% HGP-SL & 73.70{\tiny$\pm$2.69} & \second{50.33{\tiny$\pm$2.41}} & OOM & \second{55.83{\tiny$\pm$7.46}} & 73.04{\tiny$\pm$5.66} & 3 \\
% \noalign{\vskip 1.5pt}
% \hdashline
% \noalign{\vskip 1.5pt}
GRCN & 72.00{\tiny$\pm$2.12} & 49.05{\tiny$\pm$1.71} & OOM & 46.67{\tiny$\pm$7.76} & 59.82{\tiny$\pm$3.17} & 5.6 \\
GAUG & 69.30{\tiny$\pm$4.47} & 50.00{\tiny$\pm$3.56} & OOM & 51.00{\tiny$\pm$4.98} & 57.54{\tiny$\pm$9.72} & 5.6 \\ 
NeuralSparse & 72.90{\tiny$\pm$2.81} & 46.53{\tiny$\pm$4.50} & \second{81.25{\tiny$\pm$2.23}} & 51.50{\tiny$\pm$8.83} & 69.91{\tiny$\pm$4.68} & 4.2 \\ 
VIBGSL & \second{73.80{\tiny$\pm$2.66}} & 49.80{\tiny$\pm$2.53} & 76.45{\tiny$\pm$2.97} & 51.17{\tiny$\pm$8.79} & \first{74.21{\tiny$\pm$5.33}} & 3 \\
HGP-SL & 73.70{\tiny$\pm$2.69} & 50.33{\tiny$\pm$2.41} & OOM & 55.83{\tiny$\pm$7.46} & 73.04{\tiny$\pm$5.66} & 3 \\
% \noalign{\vskip 1.5pt}
% \hdashline
% \noalign{\vskip 1.5pt}
MOSGSL & \first{75.70{\tiny$\pm$2.05}} & \first{50.93{\tiny$\pm$2.82}} & \first{86.70{\tiny$\pm$4.20}} & \first{62.67{\tiny$\pm$6.50}} & \second{74.03{\tiny$\pm$4.08}} & 1.2 \\
    \bottomrule
    \end{tabular}
\end{table}

\subsection{Performance Comparison with Baselines.} We report the performance of GNNs and other algorithms using GCN as backbone in Table~\ref{tab:1}, from which we can derive several observations. Firstly, while some certain GSL methods like GRCN, GAUG, and NeuralSparse can be adapted for graph-level tasks with minor adjustments, they generally fall short compared to other methods, particularly evident on biological datasets, where they struggle to outperform the backbones (e.g., -7.5\% accuracy degradation on ENZYMES by GAUG, -4.8\% accuracy degradation on PROTEINS by GRCN). This discrepancy underscores a disparity between node and graph classification, and simply extending existing GSL paradigm to graph classification does not yield benefits. Secondly, two GSL methods specifically designed for graph classification (i.e., VIBGSL and HGP-SL) outperform GNNs and other GSL baselines in most cases, demonstrating the effectiveness of their specific designs for graph-level tasks. However, their improvements remain moderate and may have declined on specific datasets (e.g., -6.17\% accuracy degradation on ENZYMES by VIBGSL), possibly due to their inability to learn good structures under limited supervision. Thirdly, graph pooling methods generally outperform other baselines. This may result from their focus on subgraphs, demonstrating the importance of approaching graph classification from a subgraph perspective. Finally, we observe that our proposed MOSGSL consistently boosts performance with an average gain of 8.85\% over GCN. MOSGSL shows notable advantages when compared to GNNs and GSL baselines and remains competitive with graph pooling methods. Although the superiority is not significantly large over graph pooling methods, our experiments in Appx~\ref{pooling} demonstrate that MOSGSL, functioning as a data-centric GSL approach, can synergize effectively with them to further improve the performance.

We further evaluate the performance of all GSL methods when using different GNN backbones, and the results are shown in Table~\ref{tab:2}. We can observe that MOSGSL significantly outperforms other GSL baselines across all backbones. This underscores the superiority of MOSGSL as a versatile framework capable of learning better structures with various backbones.

\label{learning}
\subsection{MOSGSL under Different Learning Procedures.} Inspired by some prior work~\cite{zhou2023opengsl}, we further validate the utility of MOSGSL under different learning procedures. Specifically, we consider three distinct regimes: \textit{co-learning}, \textit{preprocessing}, and \textit{test-time augmentation}. Co-learning is a commonly used procedure where GSL methods and backbone GNN are trained together. In preprocessing, we extract the learned structures and use them to train a separate GNN from scratch. This setting aims to evaluate whether the learned structures indeed offer superiority over original ones. In test-time augmentation, given a well-trained GNN, we fix its parameters when training the GSL framework and then refine the graph structures on the test set before feeding them into the GNN, as in~\cite{ju2023graphpatcher}. 

\begin{table}[!t]
\renewcommand\arraystretch{1.05}
    \caption{Performance (\%) of all GSL models under different learning procedures. Shown is the mean~±~s.d.~of 10 runs on different folds.}
    \label{tab:3}
    \centering
    % \small
    \fontsize{9pt}{9pt}\selectfont
    \setlength\tabcolsep{6pt} % default value: 6pt
    \begin{tabular}{lcccccc}\toprule
    \textbf{Model} &\textbf{IMDB-B} &\textbf{IMDB-M} &\textbf{RDT-B} &\textbf{ENZYMES} &\textbf{PROTEINS} &\textbf{Avg. Gain} \\
    \midrule
    \rowcolor{gray!12}\multicolumn{7}{c}{\textsc{Using GCN as Backbone}} \\
    \midrule  
GCN & 72.60{\tiny$\pm$3.53} & 49.60{\tiny$\pm$1.94} & 70.60{\tiny$\pm$3.69} & 48.67{\tiny$\pm$5.71} & 66.49{\tiny$\pm$4.58} & -- \\
MOSGSL(co) & 75.50{\tiny$\pm$2.25} & 51.60{\tiny$\pm$2.05} & 82.05{\tiny$\pm$3.53} & 63.50{\tiny$\pm$5.40} & 74.03{\tiny$\pm$5.41} & 7.74 \\
MOSGSL(pre) & 74.50{\tiny$\pm$1.96} & 49.60{\tiny$\pm$2.38} & 82.50{\tiny$\pm$2.07} & 63.83{\tiny$\pm$6.62} & 73.32{\tiny$\pm$5.20} & 6.96 \\
MOSGSL(test) & 74.30{\tiny$\pm$3.09} & 50.07{\tiny$\pm$2.99} & 82.30{\tiny$\pm$1.78} & 61.67{\tiny$\pm$9.23} & 72.60{\tiny$\pm$6.26} & 6.40 \\
% \noalign{\vskip 1.5pt}
\midrule
    \rowcolor{gray!12}\multicolumn{7}{c}{\textsc{Using SAGE as Backbone}} \\
    \midrule 
% \noalign{\vskip 1.5pt}
SAGE & 72.80{\tiny$\pm$2.48} & 49.60{\tiny$\pm$2.39} & 67.80{\tiny$\pm$4.24} & 55.50{\tiny$\pm$5.11} & 68.56{\tiny$\pm$4.93} & -- \\
MOSGSL(co) & 75.90{\tiny$\pm$2.84} & 51.87{\tiny$\pm$1.65} & 79.20{\tiny$\pm$3.65} & 65.17{\tiny$\pm$5.89} & 72.42{\tiny$\pm$3.84} & 6.06 \\
MOSGSL(pre) & 74.70{\tiny$\pm$2.98} & 50.33{\tiny$\pm$3.79} & 74.65{\tiny$\pm$7.22} & 61.33{\tiny$\pm$6.23} & 72.24{\tiny$\pm$6.55} & 3.80 \\
MOSGSL(test) & 75.40{\tiny$\pm$2.91} & 49.93{\tiny$\pm$3.58} & 72.50{\tiny$\pm$7.78} & 65.67{\tiny$\pm$7.42} & 71.88{\tiny$\pm$5.58} & 4.22 \\
% \noalign{\vskip 1.5pt}
\midrule
    \rowcolor{gray!12}\multicolumn{7}{c}{\textsc{Using GIN as Backbone}} \\
    \midrule 
% \noalign{\vskip 1.5pt}
GIN & 72.00{\tiny$\pm$2.37} & 48.33{\tiny$\pm$2.05} & 82.00{\tiny$\pm$2.11} & 54.00{\tiny$\pm$5.17} & 69.18{\tiny$\pm$2.85} & -- \\
MOSGSL(co) & 75.70{\tiny$\pm$2.05} & 50.93{\tiny$\pm$2.82} & 86.70{\tiny$\pm$4.20} & 62.67{\tiny$\pm$6.50} & 74.03{\tiny$\pm$4.08} & 6.20 \\
MOSGSL(pre) & 74.50{\tiny$\pm$2.64} & 49.80{\tiny$\pm$2.39} & 87.55{\tiny$\pm$1.76} & 61.66{\tiny$\pm$4.86} & 72.96{\tiny$\pm$5.86} & 5.49 \\
MOSGSL(test) & 74.60{\tiny$\pm$2.67} & 51.67{\tiny$\pm$2.88} & 88.40{\tiny$\pm$2.09} & 57.00{\tiny$\pm$7.97} & 72.59{\tiny$\pm$5.91} & 5.05 \\
    \bottomrule
    \end{tabular}
\end{table}

% \vskip -0.185in

The results in Table~\ref{tab:3} reveal that MOSGSL performs best under the co-training regime, which is expected as co-training allows mutual adaptation between the GSL part and the backbone GNN. Under the remaining two settings, MOSGSL still demonstrates competitive performance compared to other GSL methods in Table~\ref{tab:2}. These results highlight MOSGSL's adaptability across diverse learning procedure and its potential for various practical scenarios.

% \begin{wraptable}[10]{r}{0.55\textwidth}%[!htb]
% % \fontsize{8}{9.2}\selectfont
% \vspace{-9mm}
% \caption{Ablation.}
% \vspace{2mm}
% %: standard GCN outperforms all heterophily-specific models on these graphs.}
% \setlength{\tabcolsep}{1mm}
% \begin{tabular}{lccc}
%     Method    & IMDB-B &  IMDB-M& PROTEINS\\ 
%         \toprule
% GCN     &   72.60{\tiny$\pm$3.53}      &  49.60{\tiny$\pm$1
% .94}   &66.49{\tiny$\pm$4.58}         \\
% Ours sub&73.10{\tiny$\pm$4.38} & 50.67{\tiny$\pm$3.62}&68.75{\tiny$\pm$5.58} \\
% Ours gsl&73.00{\tiny$\pm$4.12} & 49.73{\tiny$\pm$3.25}&71.84{\tiny$\pm$5.03}\\
% Ours sub+gsl&74.10{\tiny$\pm$2.81}          &  51.07{\tiny$\pm$2.14}&72.23{\tiny$\pm$7.07} \\
% Ours gsl+motif& 73.50{\tiny$\pm$2.72}   &   49.53{\tiny$\pm$3.02}&   73.12{\tiny$\pm$4.70}     \\
% Ours fixed motif & 73.50{\tiny$\pm$2.32} & 49.93{\tiny$\pm$2.58}&72.69{\tiny$\pm$5.68}\\
% MOSGSL &  75.50{\tiny$\pm$2.25}         &  51.47{\tiny$\pm$2.07}&74.03{\tiny$\pm$5.41} \\ 
% \bottomrule     
% % MLP &48.11$\pm$ 2.23 &31.68{\tiny$\pm$1.90 
% \end{tabular}
% \label{tab:pre_gcn_works_hetero} 
% \vspace{3cm}
% \end{wraptable}
\subsection{Ablation Study}

\begin{figure}[!t]
	\centering
	\begin{minipage}[!t]{0.57\linewidth}
		\centering
		\includegraphics[width=\linewidth]{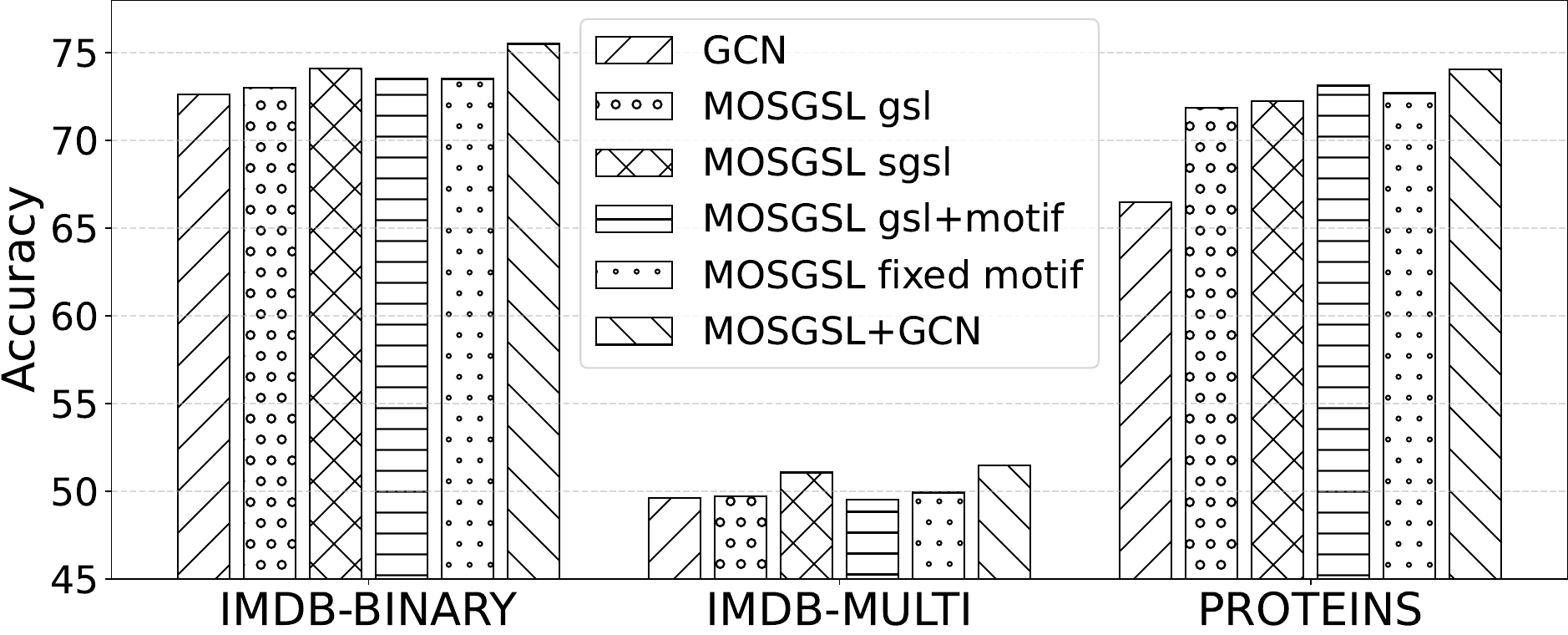}
  % \vskip -0.085in
		\caption{Performance of MOSGSL variants.}
		\label{fig:ablation}
	\end{minipage}
	% \hspace{0.02\linewidth}
        \hfill
	\begin{minipage}[!t]{0.41\linewidth}
            \vspace{-1mm}
		\centering
		\includegraphics[width=\linewidth]{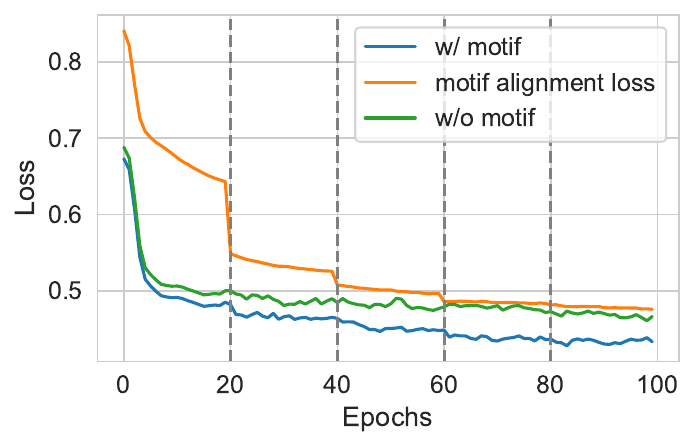}
            \vskip -0.185in
		\caption{Training dynamics.}
		\label{fig:dyna}
	\end{minipage}
	% \hspace{0.02\linewidth}
	% \begin{minipage}[t]{0.31\linewidth}
	% 	\centering
	% 	\includegraphics[width=\linewidth]{dynamic.pdf}
	% 	\caption{c}
	% 	\label{fig.viz}
	% \end{minipage}
	% \vskip -0.1in
 \vspace{-4mm}
\end{figure}

% \begin{wrapfigure}{r}{0.5\linewidth}
% \vspace{-5mm}
%     \centering
%     \includegraphics[width=\linewidth]{ablation.pdf}
%     \vspace{-7mm}
%     \caption{Ablation Studies.}
%     \label{fig:ablation}
% \end{wrapfigure}

\paragraph{Performance of MOSGSL Variants.} To accurately reflect the contributions of each component in our proposed model, we evaluated the performance of various variants of MOSGSL. Specifically, we examined 1) the removal of SGSL (gsl+motif), 2) the removal of the motif(sgsl), 3) the removal of both SGSL and motif(gsl), and 4) the fixation of the motif(fixed motif). 

% \begin{wrapfigure}{r}{0.5\linewidth}
% \vspace{-5mm}
%     \centering
%     \includegraphics[width=\linewidth]{dynamic.pdf}
%     \vspace{-8mm}
%     \caption{Training dynamics of MOSGSL}
%     \label{fig:neighborhood_dist}
% \end{wrapfigure}

As illustrated in Figure~\ref{fig:ablation}, these variants exhibited some improvements compared to the baseline, albeit not as significant as MOSGSL. Notably, among these variants, GSL (removing both SGSL and motif) exhibited relatively inferior performance, indicating the importance of both SGSL and motif. Furthermore, the performance of fixing the motif was worse than without the motif, implying the importance of motif update mechanism.

\label{dynamic}
\paragraph{Training Dynamics w/ and w/o Motif.} To further understand the internal mechanisms of MOSGSL, we plot the training dynamics of MOSGSL in Figure~\ref{fig:dyna}. It can be observed that MOSGSL achieves lower validation loss compared to the one without the guidance of motif. Moreover, with each update on motif (every 20 epochs in this case), the motif alignment loss decreases in a step-like pattern. This reveals the mutually enhancing effect between structure learning and motif extraction, where a better motif supports structure learning towards more precise refinement, which can be further leveraged to generate better motifs.

\section{Conclusion}
\label{sec:conandlimit}

In this work, we go beyond the conventional GSL paradigm and propose a novel motif-driven subgraph structure learning method MOSGSL for graph classification. By incorporating a subgraph structure learning module, MOSGSL can adaptively select important subgraphs while refining their substructures. To further capture representative structural patterns of each graph category and guide the structure learning of subgraphs, we design a motif-driven structure guidance module, where motif extraction and subgraph-motif alignment are iteratively conducted. Extensive experiments across diverse datasets validate the efficacy of our proposed MOSGSL. Furthermore, MOSGSL exhibits the benefit of being a versatile and adaptable framework capable of supporting different learning procedures and backbones. Future work includes exploring more choices for several components in MOSGSL(e.g., graph partitioning and motif extraction) and the effectiveness of MOSGSL equipped with more backbones. Additionally, it is also worth investigating what specific subgraph patterns MOSGSL has learned to make our work more interpretable.

% \section*{References}

% References follow the acknowledgments in the camera-ready paper. Use unnumbered first-level heading for
% the references. Any choice of citation style is acceptable as long as you are
% consistent. It is permissible to reduce the font size to \verb+small+ (9 point)
% when listing the references.
% Note that the Reference section does not count towards the page limit.
% \medskip

% {
% \small

% [1] Alexander, J.A.\ \& Mozer, M.C.\ (1995) Template-based algorithms for
% connectionist rule extraction. In G.\ Tesauro, D.S.\ Touretzky and T.K.\ Leen
% (eds.), {\it Advances in Neural Information Processing Systems 7},
% pp.\ 609--616. Cambridge, MA: MIT Press.

% [2] Bower, J.M.\ \& Beeman, D.\ (1995) {\it The Book of GENESIS: Exploring
%   Realistic Neural Models with the GEneral NEural SImulation System.}  New York:
% TELOS/Springer--Verlag.

% [3] Hasselmo, M.E., Schnell, E.\ \& Barkai, E.\ (1995) Dynamics of learning and
% recall at excitatory recurrent synapses and cholinergic modulation in rat
% hippocampal region CA3. {\it Journal of Neuroscience} {\bf 15}(7):5249-5262.
% }
\clearpage
\newpage

\bibliographystyle{abbrv}
{
\small
\bibliography{sample}

\begin{thebibliography}{10}

\bibitem{baek2021accurate}
J.~Baek, M.~Kang, and S.~J. Hwang.
\newblock Accurate learning of graph representations with graph multiset pooling.
\newblock In {\em The Ninth International Conference on Learning Representations}. The International Conference on Learning Representations (ICLR), 2021.

\bibitem{10.1093/bioinformatics/bti1007}
K.~M. Borgwardt, C.~S. Ong, S.~Sch\"{o}nauer, S.~V.~N. Vishwanathan, A.~J. Smola, and H.-P. Kriegel.
\newblock Protein function prediction via graph kernels.
\newblock {\em Bioinformatics}, 21(1):47–56, jan 2005.

\bibitem{bouritsas2022improving}
G.~Bouritsas, F.~Frasca, S.~Zafeiriou, and M.~M. Bronstein.
\newblock Improving graph neural network expressivity via subgraph isomorphism counting.
\newblock {\em IEEE Transactions on Pattern Analysis and Machine Intelligence}, 45(1):657--668, 2022.

\bibitem{cai2021line}
L.~Cai, J.~Li, J.~Wang, and S.~Ji.
\newblock Line graph neural networks for link prediction.
\newblock {\em IEEE Transactions on Pattern Analysis and Machine Intelligence}, 44(9):5103--5113, 2021.

\bibitem{chienadaptive}
E.~Chien, J.~Peng, P.~Li, and O.~Milenkovic.
\newblock Adaptive universal generalized pagerank graph neural network.
\newblock In {\em International Conference on Learning Representations}.

\bibitem{dai2018adversarial}
H.~Dai, H.~Li, T.~Tian, X.~Huang, L.~Wang, J.~Zhu, and L.~Song.
\newblock Adversarial attack on graph structured data.
\newblock In {\em International conference on machine learning}, pages 1115--1124. PMLR, 2018.

\bibitem{diehl2019edge}
F.~Diehl.
\newblock Edge contraction pooling for graph neural networks.
\newblock {\em arXiv preprint arXiv:1905.10990}, 2019.

\bibitem{duan2024structural}
L.~Duan, X.~Chen, W.~Liu, D.~Liu, K.~Yue, and A.~Li.
\newblock Structural entropy based graph structure learning for node classification.
\newblock In {\em Proceedings of the AAAI Conference on Artificial Intelligence}, volume~38, pages 8372--8379, 2024.

\bibitem{errica_fair_2020}
F.~Errica, M.~Podda, D.~Bacciu, and A.~Micheli.
\newblock A fair comparison of graph neural networks for graph classification.
\newblock In {\em Proceedings of the 8th {International} {Conference} on {Learning} {Representations} ({ICLR})}, 2020.

\bibitem{fatemi2023ugsl}
B.~Fatemi, S.~Abu-El-Haija, A.~Tsitsulin, M.~Kazemi, D.~Zelle, N.~Bulut, J.~Halcrow, and B.~Perozzi.
\newblock Ugsl: A unified framework for benchmarking graph structure learning.
\newblock {\em arXiv preprint arXiv:2308.10737}, 2023.

\bibitem{fatemi2021slaps}
B.~Fatemi, L.~El~Asri, and S.~M. Kazemi.
\newblock Slaps: Self-supervision improves structure learning for graph neural networks.
\newblock {\em Advances in Neural Information Processing Systems}, 34:22667--22681, 2021.

\bibitem{gao2019graph}
H.~Gao and S.~Ji.
\newblock Graph u-nets.
\newblock In {\em international conference on machine learning}, pages 2083--2092. PMLR, 2019.

\bibitem{gasteigerpredict}
J.~Gasteiger, A.~Bojchevski, and S.~G{\"u}nnemann.
\newblock Predict then propagate: Graph neural networks meet personalized pagerank.
\newblock In {\em International Conference on Learning Representations}.

\bibitem{Girvan_2002}
M.~Girvan and M.~E.~J. Newman.
\newblock Community structure in social and biological networks.
\newblock {\em Proceedings of the National Academy of Sciences}, 99(12):7821–7826, June 2002.

\bibitem{gligorijevic2021structure}
V.~Gligorijevi{\'c}, P.~D. Renfrew, T.~Kosciolek, J.~K. Leman, D.~Berenberg, T.~Vatanen, C.~Chandler, B.~C. Taylor, I.~M. Fisk, H.~Vlamakis, et~al.
\newblock Structure-based protein function prediction using graph convolutional networks.
\newblock {\em Nature communications}, 12(1):3168, 2021.

\bibitem{0eb15824-3d89-3bf3-8d79-2cf60ef065fa}
M.~Granovetter.
\newblock The strength of weak ties: A network theory revisited.
\newblock {\em Sociological Theory}, 1:201--233, 1983.

\bibitem{graphsage}
W.~L. Hamilton, R.~Ying, and J.~Leskovec.
\newblock Inductive representation learning on large graphs.
\newblock In {\em Proceedings of the 31st International Conference on Neural Information Processing Systems}, NIPS'17, page 1025–1035, Red Hook, NY, USA, 2017. Curran Associates Inc.

\bibitem{in2024self}
Y.~In, K.~Yoon, K.~Kim, K.~Shin, and C.~Park.
\newblock Self-guided robust graph structure refinement.
\newblock {\em arXiv preprint arXiv:2402.11837}, 2024.

\bibitem{jin2020graph}
W.~Jin, Y.~Ma, X.~Liu, X.~Tang, S.~Wang, and J.~Tang.
\newblock Graph structure learning for robust graph neural networks.
\newblock In {\em Proceedings of the 26th ACM SIGKDD international conference on knowledge discovery \& data mining}, pages 66--74, 2020.

\bibitem{ju2023graphpatcher}
M.~Ju, T.~Zhao, W.~Yu, N.~Shah, and Y.~Ye.
\newblock Graphpatcher: Mitigating degree bias for graph neural networks via test-time augmentation.
\newblock In {\em Thirty-seventh Conference on Neural Information Processing Systems}, 2023.

\bibitem{kipf2017semisupervised}
T.~N. Kipf and M.~Welling.
\newblock Semi-supervised classification with graph convolutional networks.
\newblock In {\em International Conference on Learning Representations}, 2017.

\bibitem{10.1145/3357384.3357880}
J.~B. Lee, R.~A. Rossi, X.~Kong, S.~Kim, E.~Koh, and A.~Rao.
\newblock Graph convolutional networks with motif-based attention.
\newblock CIKM '19, page 499–508, New York, NY, USA, 2019. Association for Computing Machinery.

\bibitem{li2022reliable}
K.~Li, Y.~Liu, X.~Ao, J.~Chi, J.~Feng, H.~Yang, and Q.~He.
\newblock Reliable representations make a stronger defender: Unsupervised structure refinement for robust gnn.
\newblock In {\em Proceedings of the 28th ACM SIGKDD Conference on Knowledge Discovery and Data Mining}, pages 925--935, 2022.

\bibitem{li_gslb_2023}
Z.~Li, X.~Sun, Y.~Luo, Y.~Zhu, D.~Chen, Y.~Luo, X.~Zhou, Q.~Liu, S.~Wu, L.~Wang, and J.~Yu.
\newblock {GSLB}: {The} {Graph} {Structure} {Learning} {Benchmark}.
\newblock In A.~Oh, T.~Naumann, A.~Globerson, K.~Saenko, M.~Hardt, and S.~Levine, editors, {\em Advances in {Neural} {Information} {Processing} {Systems}}, volume~36, pages 30306--30318. Curran Associates, Inc., 2023.

\bibitem{liu2022towards}
Y.~Liu, Y.~Zheng, D.~Zhang, H.~Chen, H.~Peng, and S.~Pan.
\newblock Towards unsupervised deep graph structure learning.
\newblock In {\em Proceedings of the ACM Web Conference 2022}, pages 1392--1403, 2022.

\bibitem{ijcai2023p449}
G.~Ma, C.~Hu, L.~Ge, and H.~Zhang.
\newblock Multi-view robust graph representation learning for graph classification.
\newblock In E.~Elkind, editor, {\em Proceedings of the Thirty-Second International Joint Conference on Artificial Intelligence, {IJCAI-23}}, pages 4037--4045. International Joint Conferences on Artificial Intelligence Organization, 8 2023.
\newblock Main Track.

\bibitem{morris2020tudataset}
C.~Morris, N.~M. Kriege, F.~Bause, K.~Kersting, P.~Mutzel, and M.~Neumann.
\newblock Tudataset: A collection of benchmark datasets for learning with graphs, 2020.

\bibitem{peng2020motif}
H.~Peng, J.~Li, Q.~Gong, Y.~Ning, S.~Wang, and L.~He.
\newblock Motif-matching based subgraph-level attentional convolutional network for graph classification.
\newblock In {\em Proceedings of the AAAI conference on artificial intelligence}, volume~34, pages 5387--5394, 2020.

\bibitem{10.1145/3038912.3052588}
M.~Shao, J.~Li, F.~Chen, H.~Huang, S.~Zhang, and X.~Chen.
\newblock An efficient approach to event detection and forecasting in dynamic multivariate social media networks.
\newblock In {\em Proceedings of the 26th International Conference on World Wide Web}, WWW '17, page 1631–1639, Republic and Canton of Geneva, CHE, 2017. International World Wide Web Conferences Steering Committee.

\bibitem{sun2022graph}
Q.~Sun, J.~Li, H.~Peng, J.~Wu, X.~Fu, C.~Ji, and S.~Y. Philip.
\newblock Graph structure learning with variational information bottleneck.
\newblock In {\em Proceedings of the AAAI Conference on Artificial Intelligence}, volume~36, pages 4165--4174, 2022.

\bibitem{sun2021sugar}
Q.~Sun, J.~Li, H.~Peng, J.~Wu, Y.~Ning, P.~S. Yu, and L.~He.
\newblock Sugar: Subgraph neural network with reinforcement pooling and self-supervised mutual information mechanism.
\newblock In {\em Proceedings of the Web Conference 2021}, pages 2081--2091, 2021.

\bibitem{velivckovicgraph}
P.~Veli{\v{c}}kovi{\'c}, G.~Cucurull, A.~Casanova, A.~Romero, P.~Li{\`o}, and Y.~Bengio.
\newblock Graph attention networks.
\newblock In {\em International Conference on Learning Representations}.

\bibitem{wang2023prose}
H.~Wang, Y.~Fu, T.~Yu, L.~Hu, W.~Jiang, and S.~Pu.
\newblock Prose: Graph structure learning via progressive strategy.
\newblock In {\em Proceedings of the 29th ACM SIGKDD Conference on Knowledge Discovery and Data Mining}, pages 2337--2348, 2023.

\bibitem{wang2021graph}
R.~Wang, S.~Mou, X.~Wang, W.~Xiao, Q.~Ju, C.~Shi, and X.~Xie.
\newblock Graph structure estimation neural networks.
\newblock In {\em Proceedings of the Web Conference 2021}, pages 342--353, 2021.

\bibitem{wu2020comprehensive}
Z.~Wu, S.~Pan, F.~Chen, G.~Long, C.~Zhang, and S.~Y. Philip.
\newblock A comprehensive survey on graph neural networks.
\newblock {\em IEEE transactions on neural networks and learning systems}, 32(1):4--24, 2020.

\bibitem{xupowerful}
K.~Xu, W.~Hu, J.~Leskovec, and S.~Jegelka.
\newblock How powerful are graph neural networks?
\newblock In {\em International Conference on Learning Representations}.

\bibitem{xu2018representation}
K.~Xu, C.~Li, Y.~Tian, T.~Sonobe, K.-i. Kawarabayashi, and S.~Jegelka.
\newblock Representation learning on graphs with jumping knowledge networks.
\newblock In {\em International conference on machine learning}, pages 5453--5462. PMLR, 2018.

\bibitem{8924759}
Q.~Xuan, J.~Wang, M.~Zhao, J.~Yuan, C.~Fu, Z.~Ruan, and G.~Chen.
\newblock Subgraph networks with application to structural feature space expansion.
\newblock {\em IEEE Transactions on Knowledge and Data Engineering}, 33(6):2776--2789, 2021.

\bibitem{you2020graph}
Y.~You, T.~Chen, Y.~Sui, T.~Chen, Z.~Wang, and Y.~Shen.
\newblock Graph contrastive learning with augmentations.
\newblock {\em Advances in neural information processing systems}, 33:5812--5823, 2020.

\bibitem{yu2021graph}
D.~Yu, R.~Zhang, Z.~Jiang, Y.~Wu, and Y.~Yang.
\newblock Graph-revised convolutional network.
\newblock In {\em Machine Learning and Knowledge Discovery in Databases: European Conference, ECML PKDD 2020, Ghent, Belgium, September 14--18, 2020, Proceedings, Part III}, pages 378--393. Springer, 2021.

\bibitem{xia2024gnncert}
zaishuo xia, H.~Yang, B.~Wang, and J.~Jia.
\newblock {GNNC}ert: Deterministic certification of graph neural networks against adversarial perturbations.
\newblock In {\em The Twelfth International Conference on Learning Representations}, 2024.

\bibitem{zha2023data-centric-perspectives}
D.~Zha, Z.~P. Bhat, K.-H. Lai, F.~Yang, and X.~Hu.
\newblock Data-centric ai: Perspectives and challenges.
\newblock In {\em SDM}, 2023.

\bibitem{zhang2018link}
M.~Zhang and Y.~Chen.
\newblock Link prediction based on graph neural networks.
\newblock {\em Advances in neural information processing systems}, 31, 2018.

\bibitem{10429945}
S.~Zhang, Z.~Hu, A.~Subramonian, and Y.~Sun.
\newblock Motif-driven contrastive learning of graph representations.
\newblock {\em IEEE Transactions on Knowledge and Data Engineering}, pages 1--12, 2024.

\bibitem{zhang2019hierarchical}
Z.~Zhang, J.~Bu, M.~Ester, J.~Zhang, C.~Yao, Z.~Yu, and C.~Wang.
\newblock Hierarchical graph pooling with structure learning.
\newblock {\em arXiv preprint arXiv:1911.05954}, 2019.

\bibitem{zhang2022protgnn}
Z.~Zhang, Q.~Liu, H.~Wang, C.~Lu, and C.~Lee.
\newblock Protgnn: Towards self-explaining graph neural networks.
\newblock In {\em Proceedings of the AAAI Conference on Artificial Intelligence}, volume~36, pages 9127--9135, 2022.

\bibitem{zhang2021motif}
Z.~Zhang, Q.~Liu, H.~Wang, C.~Lu, and C.-K. Lee.
\newblock Motif-based graph self-supervised learning for molecular property prediction.
\newblock {\em Advances in Neural Information Processing Systems}, 34:15870--15882, 2021.

\bibitem{zhao2023self}
J.~Zhao, Q.~Wen, M.~Ju, C.~Zhang, and Y.~Ye.
\newblock Self-supervised graph structure refinement for graph neural networks.
\newblock In {\em Proceedings of the Sixteenth ACM International Conference on Web Search and Data Mining}, pages 159--167, 2023.

\bibitem{zhao2021data}
T.~Zhao, Y.~Liu, L.~Neves, O.~Woodford, M.~Jiang, and N.~Shah.
\newblock Data augmentation for graph neural networks.
\newblock In {\em Proceedings of the aaai conference on artificial intelligence}, volume~35, pages 11015--11023, 2021.

\bibitem{zheng2020robust}
C.~Zheng, B.~Zong, W.~Cheng, D.~Song, J.~Ni, W.~Yu, H.~Chen, and W.~Wang.
\newblock Robust graph representation learning via neural sparsification.
\newblock In {\em International Conference on Machine Learning}, pages 11458--11468. PMLR, 2020.

\bibitem{zhou2023opengsl}
Z.~Zhou, S.~Zhou, B.~Mao, X.~Zhou, J.~Chen, Q.~Tan, D.~Zha, Y.~Feng, C.~Chen, and C.~Wang.
\newblock Opengsl: A comprehensive benchmark for graph structure learning.
\newblock In {\em Thirty-seventh Conference on Neural Information Processing Systems Datasets and Benchmarks Track}, 2023.

\bibitem{zhu2019robust}
D.~Zhu, Z.~Zhang, P.~Cui, and W.~Zhu.
\newblock Robust graph convolutional networks against adversarial attacks.
\newblock In {\em Proceedings of the 25th ACM SIGKDD international conference on knowledge discovery \& data mining}, pages 1399--1407, 2019.

\bibitem{zou2023se}
D.~Zou, H.~Peng, X.~Huang, R.~Yang, J.~Li, J.~Wu, C.~Liu, and P.~S. Yu.
\newblock Se-gsl: A general and effective graph structure learning framework through structural entropy optimization.
\newblock In {\em Proceedings of the ACM Web Conference 2023}, pages 499--510, 2023.

\bibitem{zugner_adversarial_2019}
D.~Z{\"u}gner and S.~G{\"u}nnemann.
\newblock Adversarial attacks on graph neural networks via meta learning.
\newblock In {\em International Conference on Learning Representations (ICLR)}, 2019.

\end{thebibliography}
}

%%%%%%%%%%%%%%%%%%%%%%%%%%%%%%%%%%%%%%%%%%%%%%%%%%%%%%%%%%%%
\clearpage
\newpage

\appendix

\section{More Discussion w.r.t related works}

Exploiting subgraph information in GNNs is not new. As local substructures in a graph always contain vital characteristics and prominent patterns, a large body of works has been proposed to fully exploit the semantics of subgraphs, including motif-based methods~\cite{10.1145/3357384.3357880, peng2020motif, zhang2021motif}, subgraph isomorphism counting~\cite{bouritsas2022improving}, and rule-based extraction~\cite{8924759}. The key distinction between our work and theirs lies in that we aim to address the sub-optimality of the original structure from a data-centric perspective and learn a broadly effective structure for downstream tasks. Additionally, we extract motifs in an automatic manner, as opposed to requiring prior knowledge as in~\cite{zhang2021motif}. The idea of automatically extracting motifs also appears in~\cite{10429945}, but with completely different implementations compared to ours and without considering structure learning.

The motif in our approach plays a role similar to the anchor in SUBLIME~\cite{liu2022towards}. However, the difference lies in that SUBLIME utilizes the original structure as guidance and belongs to the conventional GSL paradigm, while we adopt subgraph structure learning and extract common structural motifs as guidance in a label-aware manner. Another work~\cite{zhang2022protgnn} also contains a similar concept close to our motif (defined as prototypes within it). Our method differs from it in that we use extracted motifs to guide structure learning rather than directly predicting graph labels. Furthermore, MOSGSL allows different subgraphs within a graph to have different patterns and employs a completely different approach to extracting motifs compared with~\cite{zhang2022protgnn}.

\section{Datasets}
\label{dataset}

We use five datasets from~\cite{morris2020tudataset}, which have been widely utilized by previous works~\cite{errica_fair_2020, sun2022graph, sun2021sugar}. The included datasets consist of social network datasets (IMDB-BINARY, IMDB-MULTI, and REDDIT-BINARY) and bioinformatics datasets (ENZYMES, PROTEINS). IMDB-BINARY and IMDB-MULTI are movie collaboration datasets, where nodes represent actors/actresses and edges represent their collaborations. Each graph corresponds to movies of a specific genre, and the task is to predict the genre of each graph. REDDIT-BINARY is a dataset created from a network discussion platform. In each graph, nodes correspond to users, and there is an edge between two nodes if one user replies to another. The community to which each graph belongs is the label to be predicted. ENZYMES is a dataset of protein tertiary structures, where the label of each graph is one of the 6 EC top-level classes. In PROTEINS, nodes are secondary structure elements (SSEs), and there is an edge between two nodes if they are neighbors in the amino-acid sequence or in 3D space. The task is to predict whether each graph is a non-enzyme. 

The detailed statistics of these datasets are listed in the table.

\begin{table}[h]
    \caption{Statistics of the datasets used.}
    \label{tab:datasets}
    \setlength\tabcolsep{4pt} % default value: 6pt
    \centering
    \renewcommand\arraystretch{1}
    \resizebox{0.8\textwidth}{!}{
    \begin{tabular}{l|ccccc}
    \toprule
    \textbf{Dataset}&\textbf{Category}& \textbf{Graphs} & \textbf{Classes} & \textbf{Avg nodes} & \textbf{Avg edges} \\
    \midrule
    IMDB-B & Social & 1000 & 2 & 19.8 & 96.5\\
    IMDB-M & Social & 1500 & 3 & 13.0 & 65.9\\
    RDT-B & Social & 2000 & 2 & 429.6 & 497.8\\
    ENZYMES & Protein & 600 & 6 & 32.6 & 124.2\\
    PROTEINS & Protein & 1113 & 2 & 39.0 & 72.8\\
    \bottomrule
    \end{tabular}
    }
\label{table:dataset}
\end{table}

\section{Experimental Details}
\label{sec:appdix_experiment}

\subsection{Hardware and Software Configuration}
Our experiments are mostly conducted on a Linux server with an Intel(R) Xeon(R) Silver 4216 CPU @ 2.10GHz, 125 GB RAM, and an NVIDIA GTX 2080 Ti GPU (12GB). We implement our method using PyTorch 2.0.1 and PyG 2.5.0. For GSL baselines we use the reproduced versions from either OpenGSL~\cite{zhou2023opengsl} or GSLB~\cite{li_gslb_2023}, two recently released libraries on GSL. For other graph pooling methods we use the implementations in PyG. These code repositories are listed as follows. 

\begin{itemize}[leftmargin=*]
    \item \url{https://github.com/OpenGSL/OpenGSL}
    \item \url{https://github.com/GSL-Benchmark/GSLB}
    \item \url{https://github.com/pyg-team/pytorch_geometric/tree/master/torch_geometric/nn/pool.}
\end{itemize}

It is worth noting that the performances of GNN and GSL baselines in our experiments are similar to the results in GSLB~\cite{li_gslb_2023}.

\subsection{Settings}

In the experiments, we employ 10-fold cross-validation and present the average accuracy along with the standard deviation. BatchNorm is incorporated into all models, and global average pooling is applied to both GNN and GSL methods. To ensure fairness, the number of GNN backbone layers in all algorithms is set at 2, with a fixed dropout rate of 0.5.

\subsection{Hyperparameter}

\label{hyper}

For all models, the hyperparameters to be tuned include the learning rate ($lr\in \{0.1,0.01,0.001\}$), weight decay ($wd \in \{5\mathrm{e}{-4},5\mathrm{e}{−5},5\mathrm{e}{−6},0\}$), and batch size ($bs\in \{32,64,128\}$). For other hyperparameters of the baseline models, we follow their original settings.

The key hyperparameters we tune for GRAPHPATCHER include the number of motifs specified for each category $R \in \{1,2,3,4,5\}$, the ratio of selected subgraphs $\epsilon \in \{0.2,0.4,0.6,0.8\}$ during motif update, the coefficient of the motif loss $\lambda \in \{0.5, 0.2,0.1,0.05, 0.01\}$, and the structural fusion coefficient $\gamma \in \{0.1,0.3,0.5,0.7\}$. Detailed hyperparameter configurations for all datasets when using GCN as the backbone are also listed in table. More comprehensive hyperparameter settings can be found in our publicly available code repository.

\begin{table}[h]
    \centering
    \caption{Hyper-parameters used for.}
    \resizebox{1\columnwidth}{!}{
    \begin{tabular}{lccccc}
    \toprule 
    \textbf{Hyper-param. }  & \textbf{IMDB-B} & \textbf{IMDB-M} & \textbf{RDT-B} & \textbf{ENZYMES} & \textbf{PROTEINS} \\
    \cmidrule(r){1-6} 
    Number of motifs $R$ & 2 & 4 & 2 & 2 & 1 \\
    Ratio $\epsilon$ of filtered subgraphs & 0.6 & 0.6 & 0.4 & 0.4 & 0.6  \\
    Coefficient $\lambda$ of the motif loss & 0.1 & 0.01 & 0.1 & 0.01 & 0.01  \\
    Structural fusion coefficient $\gamma$ & 0.5 & 0.3 & 0.3 & 0.3 & 0.5  \\
    Batch size & 64 & 128 & 128 & 128 & 32 \\
    Learning rate & 1e-3 & 1e-2 & 1e-2 & 1e-3 & 1e-3 \\
    Weight decay & 5e-4 & 5e-4 & 0 & 5e-4 & 5e-6 \\
    \bottomrule 
    \end{tabular}}
    \label{tab:hyper}
\end{table}

Additionally, we provide an analysis of some hyperparameters in Appx~\ref{hyperexp}.

\section{Additional Results}

\subsection{Additional Results on Initialization Methods of Motif}
\label{motif}

We consider two ways of initializing motifs: 1) random initialization. 2) initialization from pretrained subgraph representations. Specifically, the second initialization method refers to pretraining a simple Subgraph GNN, which takes subgraphs as input and simply adds up the subgraph representation as graph representations to predict labels. We use the trained GNN as an encoder to obtain representations for all subgraphs, and then use the clustering method described in Sec~\ref{sec31} to get initial motif representations.

Table~\ref{tab:compa} shows the performance of the two initialization methods. It is evident that initializing motifs from pretrained subgraph representations performs better. This is as expected, as the second method provides more meaningful motif representations as start points. Nevertheless, MOSGSL using the first method still shows improvement compared to the backbone.

\begin{table}[h]
\vspace{-5mm}
\renewcommand\arraystretch{1.05}
    \caption{Results on initialization methods of motif}
    \label{tab:compa}
    \centering
    % \small
    \fontsize{9pt}{9pt}\selectfont
    \setlength\tabcolsep{6pt} % default value: 6pt
    \begin{tabular}{lccc}\toprule
    \textbf{Initialization} &\textbf{IMDB-B} &\textbf{IMDB-M} &\textbf{PROTEINS} \\
    \midrule 
        Pretrain & 75.5 & 51.60 & 73.80\\
        Random & 74.9 & 50.80 & 72.06\\
    \bottomrule
    \end{tabular}
\end{table}

\subsection{Compatibility with Graph Pooling Methods}

\begin{wraptable}[10]{r}{0.55\textwidth}
\vspace{-5mm}
\renewcommand\arraystretch{1.05}
    \caption{Results.}
    \label{tab:compa1}
    \centering
    % \small
    \fontsize{9pt}{9pt}\selectfont
    \setlength\tabcolsep{6pt} % default value: 6pt
    \begin{tabular}{lccc}\toprule
    \textbf{Model} &\textbf{IMDB-B} &\textbf{IMDB-M} &\textbf{PROTEINS} \\
    \midrule 
GMT & 73.30& 50.07 & 73.77\\
+MOSGSL & 74.70 & 50.33 & 75.02\\
\midrule 
TopKPool & 74.70& 49.93 & 71.33\\
+MOSGSL & 74.90 & 50.53 & 74.53\\
\midrule 
EdgePool & 73.40& 50.80 & 71.34\\
+MOSGSL & 74.70 & 51.00 & 73.23\\
    \bottomrule
    \end{tabular}
\end{wraptable}

\label{pooling}
Here we investigate the compatibility of MOSGSL with advanced models beyond basic GNNs. Specifically, we assess the performance of three graph pooling models trained on the structures acquired through MOSGSL. Notably, we employ a preprocessing regime for efficiency, where MOSGSL undergoes pre-training using a GCN backbone, which is afterwards substituted by these models. The results in Table~\ref{tab:compa} indicate that the structures learned by MOSGSL seamlessly integrate with these models, leading to further enhancements. These enhancements are expected to be more pronounced with the adoption of a co-learning strategy, as discussed in Section~\ref{learning}. We defer this exploration to future research, along with investigating the compatibility of MOSGSL with additional methodologies.

\subsection{Additional Results on Hyperparameters}
\label{hyperexp}

\begin{figure}[!h]
	\centering
	\begin{minipage}[t]{0.41\linewidth}
		\centering
		\includegraphics[width=\linewidth]{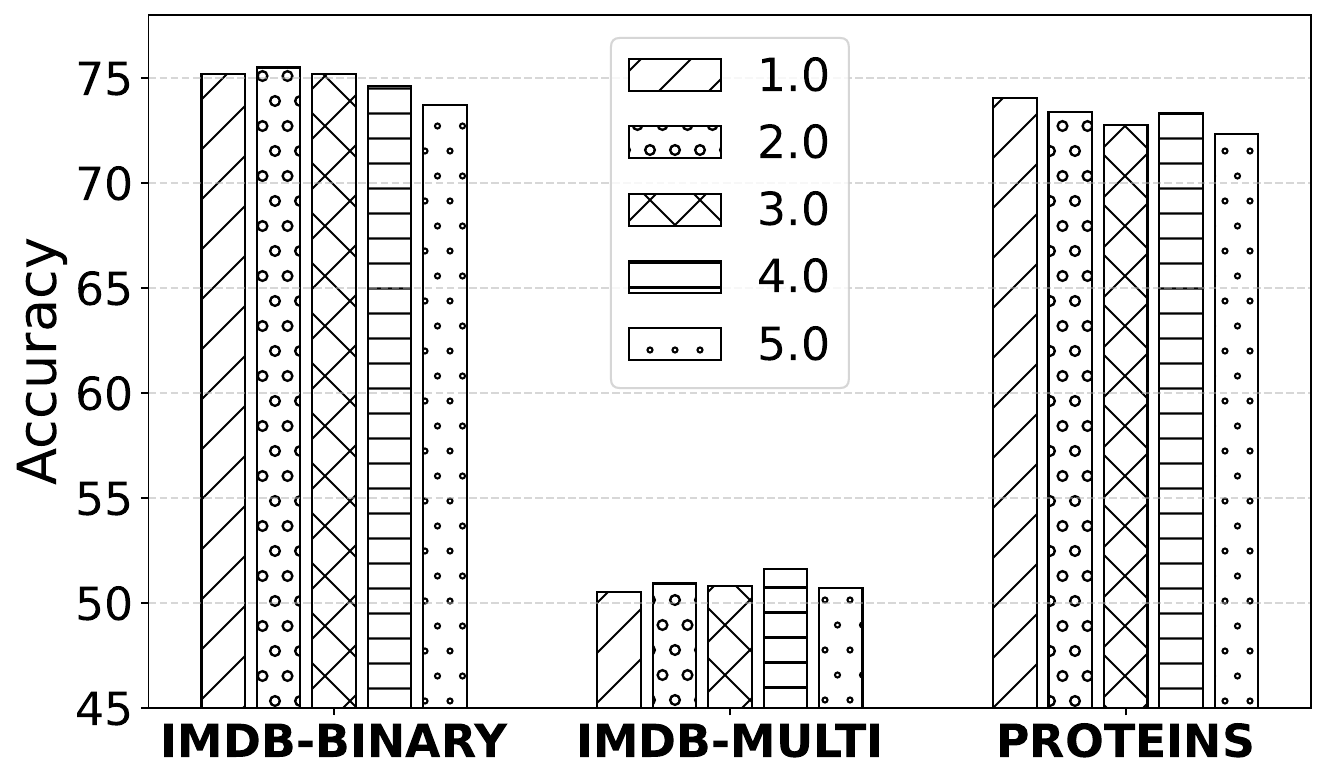}
		\caption{Number of motifs per class $R$.}
		\label{fig.5}
	\end{minipage}
	\hspace{0.02\linewidth}
	\begin{minipage}[t]{0.41\linewidth}
		\centering
		\includegraphics[width=\linewidth]{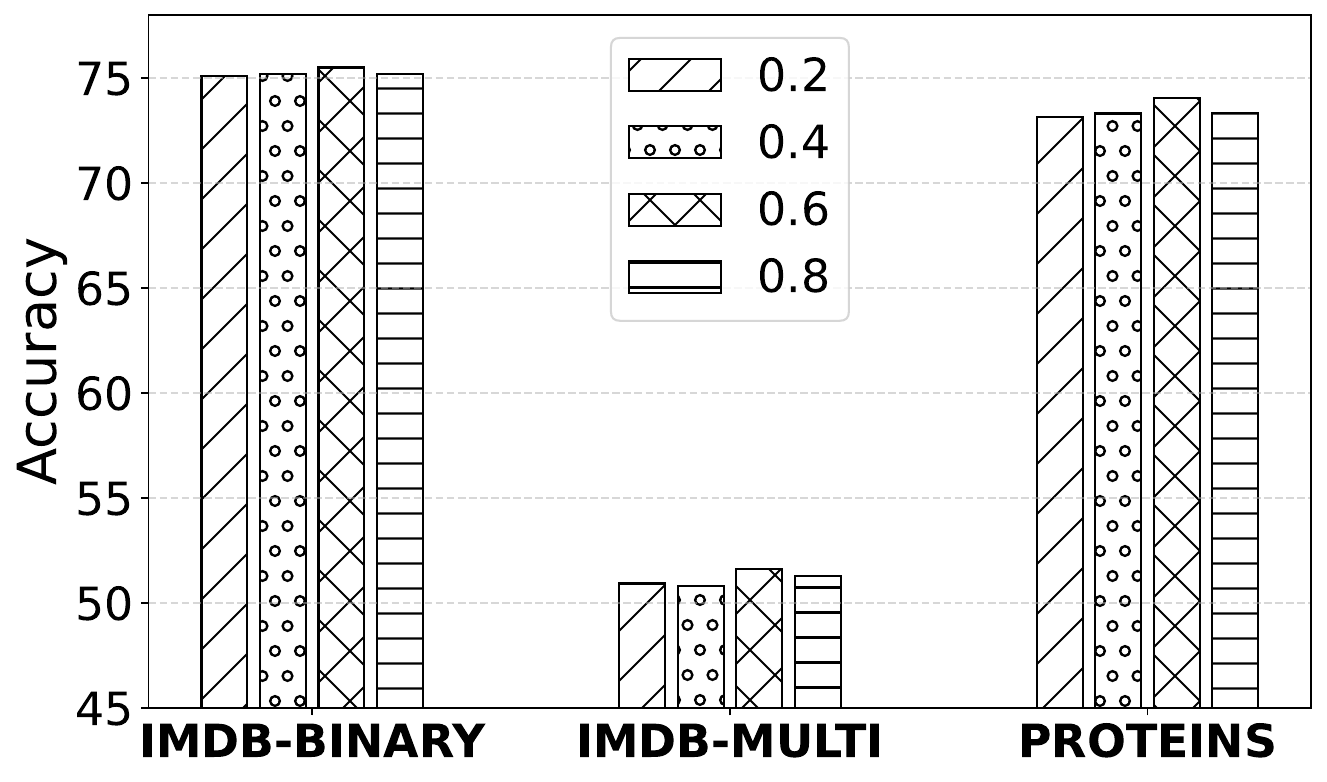}
		\caption{$\epsilon$ during motif update.}
		\label{fig.6}
	\end{minipage}
	% \hspace{0.02\linewidth}
	% \begin{minipage}[t]{0.31\linewidth}
	% 	\centering
	% 	\includegraphics[width=\linewidth]{dynamic.pdf}
	% 	\caption{c}
	% 	\label{fig.viz}
	% \end{minipage}
	% \vskip -0.1in
        \centering
	\begin{minipage}[t]{0.41\linewidth}
		\centering
		\includegraphics[width=\linewidth]{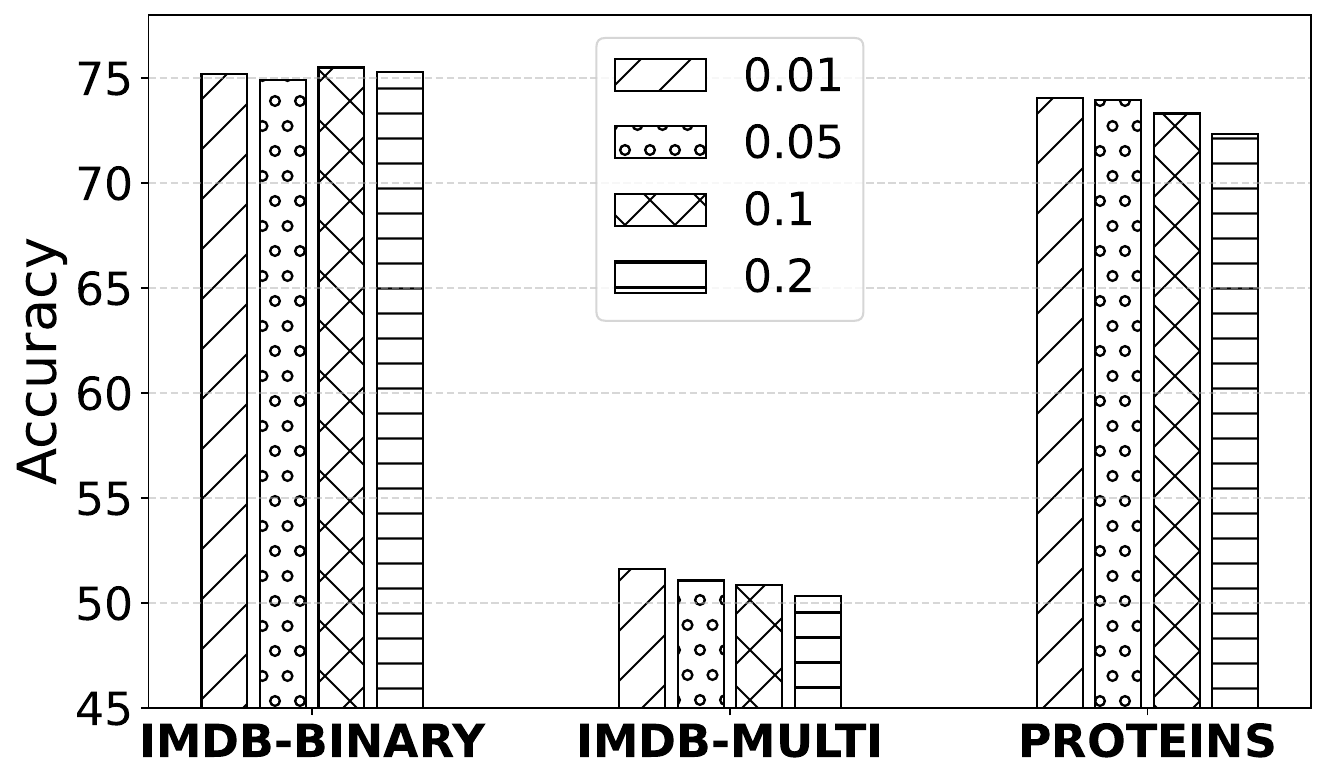}
		\caption{Number of motifs per class $R$.}
		\label{fig.7}
	\end{minipage}
        \hspace{0.02\linewidth}
	\begin{minipage}[t]{0.41\linewidth}
		\centering
		\includegraphics[width=\linewidth]{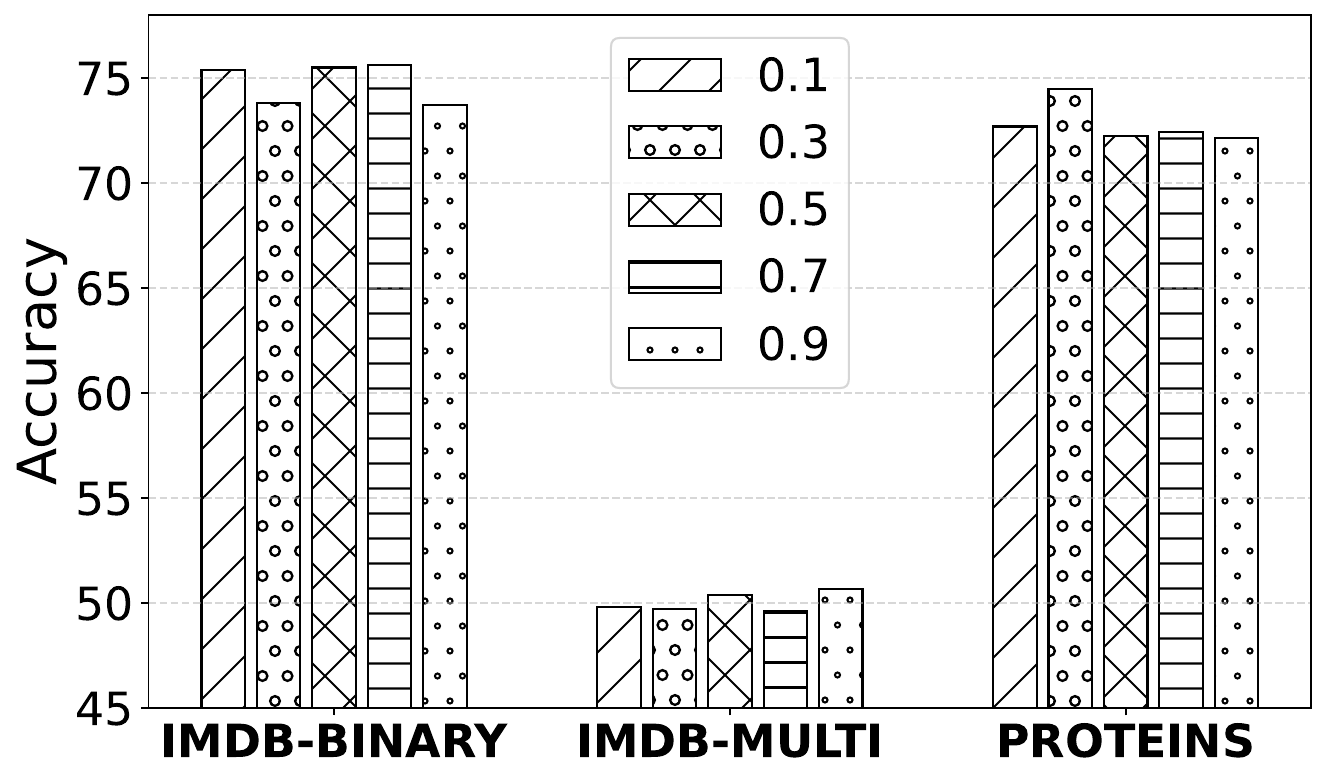}
		\caption{structural fusion coefficient $\gamma$}
		\label{fig.8}
	\end{minipage}
\end{figure}

Here we present MOSGSL's performance variations with respect to hyperparameters and analyze the effects of two hyperparameters: the number of motifs per class $R=L/||\mathcal{Y}||$ and the proportion $\epsilon$ of subgraphs filtered for updating motifs. The results are presented in Figure~\ref{fig.5} and Figure\ref{fig.6}. The performance shows minor fluctuations but consistently surpasses the backbone model. Specifically, for $R$, different datasets exhibit varying preferences (2 for IMDB-B, 4 for IMDB-M, 1 for PROTEINS), reflecting their unique structural characteristics. For $\epsilon$, a moderate value 0.6 is favored, capturing most relevant subgraphs while excluding unreliable ones. Larger values of $\epsilon$ may introduce more noise, while smaller values may lead to inadequate representation of class-specific structural patterns by the motifs.

\section{Broader Impacts}
Our work has several positive social impacts. First, our approach is a flexible framework that can adapt to various backbones and learning procedures. Others can easily utilize our framework for their respective graph-level machine learning tasks. Besides, as a data-centric GSL method, the structures we learn can function independently as better structures compared to the original structures. This could be beneficial in certain real-world scenarios.
\label{sec:app_impact}

\end{document}